\documentclass[letterpaper, 10 pt, conference]{ieeeconf}  

\IEEEoverridecommandlockouts                              

\usepackage[linesnumbered,ruled,vlined]{algorithm2e}
\usepackage{algorithmic}
\usepackage{amsmath,amsfonts,amssymb}
\usepackage{graphicx}
\graphicspath{{figures/}}
\usepackage{epsfig} 
\usepackage{times}
\usepackage{siunitx}
\usepackage{hyperref}
\usepackage{booktabs}
\usepackage[capitalise]{cleveref}
\let\labelindent\relax
\usepackage[inline]{enumitem}
\usepackage{multirow}
\usepackage[normalem]{ulem}
\Crefname{algorithm}{Alg.}{Algs.}
\Crefname{section}{Sec.}{Secs.}
\Crefname{equation}{Eq.}{Eqs.}
\usepackage{caption}
\usepackage{authblk}
\usepackage{booktabs}
\usepackage{comment}
\usepackage{todonotes}

\usepackage{comment}
\usepackage{tikz}
\usetikzlibrary{positioning}
\usepackage{xcolor}

\newcommand{\timelabelfill}{yellow!22} 

\tikzset{
  methodbadge/.style={
    font=\small\bfseries, fill=gray!15, draw=black!40,
    rounded corners=2pt, inner sep=2pt, text=black
  },
  timebadge/.style={
    font=\small, fill=\timelabelfill, draw=none,   
    rounded corners=2pt, inner sep=1.2pt, text=black
  },
  collisionbadge/.style={
    font=\footnotesize\bfseries,
    text=red!75!black,   
    fill=none,           
    draw=black!40            
  }  
}

\newcommand{\tilewidth}{.18\linewidth}

\newcommand{\tileTop}[3]{
\begin{tikzpicture}[baseline=(img.base)]
  \node[inner sep=0] (img) {\includegraphics[width=\tilewidth]{#1}};
  \node[anchor=north, yshift=2pt, methodbadge] at (img.north) {#2};
  \node[anchor=south west, xshift=2pt, yshift=2pt, timebadge] at (img.south west) {#3};
\end{tikzpicture}%
}

\newcommand{\tile}[2]{
\begin{tikzpicture}[baseline=(img.base)]
  \node[inner sep=0] (img) {\includegraphics[width=\tilewidth]{#1}};
  \node[anchor=south west, xshift=2pt, yshift=2pt, timebadge] at (img.south west) {#2};
\end{tikzpicture}%
}

\newcommand{\tileTopC}[4]{%
\begin{tikzpicture}[baseline=(img.base)]
  \node[inner sep=0] (img) {\includegraphics[width=\tilewidth]{#1}};
  \node[anchor=north, yshift=2pt, methodbadge] at (img.north) {#2};
  \node[anchor=south west, xshift=2pt, yshift=2pt, timebadge] at (img.south west) {#3};
  \ifx\relax#4\relax\else
    \node[anchor=south east, xshift=-2pt, yshift=2pt, collisionbadge] at (img.south east) {#4};
  \fi
\end{tikzpicture}%
}

\newcommand{\tileC}[3]{%
\begin{tikzpicture}[baseline=(img.base)]
  \node[inner sep=0] (img) {\includegraphics[width=\tilewidth]{#1}};
  \node[anchor=south west, xshift=2pt, yshift=2pt, timebadge] at (img.south west) {#2};
  \ifx\relax#3\relax\else
    \node[anchor=south east, xshift=-2pt, yshift=2pt, collisionbadge] at (img.south east) {#3};
  \fi
\end{tikzpicture}%
}

\usepackage[
    sorting=none,
    giveninits=true,
    maxbibnames=9,
    maxcitenames=2,backend=biber
    ]{biblatex}
\renewcommand{\baselinestretch}{0.999}

\title{\LARGE \bf
HiCrowd: Hierarchical Crowd Flow Alignment for Dense Human Environments

}

\author{
Yufei Zhu$^{1}$, Shih-Min Yang$^{1}$, Martin Magnusson$^{1}$, Allan Wang$^{2}$%
\thanks{$^{1}$Robot Navigation and Perception Lab, AASS Research Center, 
{\"O}rebro University, Sweden {\tt\small yufei.zhu@oru.se}}%
\thanks{$^{2}$Miraikan -- The National Museum of Emerging Science and Innovation, Japan. This work received funding from the European Union’s Horizon 2020 research and innovation programme under grant agreement No 101070596 (euRobin).}%
}

\begin{document}

\maketitle
\thispagestyle{empty}
\pagestyle{empty}

\begin{abstract}

Navigating through dense human crowds remains a significant challenge for mobile robots. A key issue is the freezing robot problem, where the robot struggles to find safe motions and becomes stuck within the crowd.
To address this, we propose HiCrowd, a hierarchical framework that integrates reinforcement learning (RL) with model predictive control (MPC). HiCrowd leverages surrounding pedestrian motion as guidance, enabling the robot to align with compatible crowd flows. A high-level RL policy generates a follow point to align the robot with a suitable pedestrian group, while a low-level MPC safely tracks this guidance with short horizon planning. The method combines long-term crowd aware decision making with safe short-term execution. 
We evaluate HiCrowd against reactive and learning-based baselines in offline setting (replaying recorded human trajectories) and online setting (human trajectories are updated to react to the robot in simulation). Experiments on a real-world dataset and a synthetic crowd dataset show that our method outperforms in navigation efficiency and safety, while reducing freezing behaviors. We further validate through real-world deployment in a public museum and Expo 2025 Osaka, where it navigates dense pedestrian flows without retraining, demonstrating robust and socially aware behavior.
Our results suggest that leveraging human motion as guidance, rather than treating humans solely as dynamic obstacles, provides a powerful principle for safe and efficient robot navigation in crowds. Project code and demos are available at \url{https://github.com/test-bai-cpu/HiCrowd}.
\end{abstract}




\section{Introduction}
\label{sec:introduction}

With rising cases of robots navigating through dense human environments, mobile robots must balance efficiency, safety, and social compliance~\cite{triebel2016spencer}. A key challenge in this context is the \emph{freezing robot problem} (FRP)~\cite{trautman2010unfreezing}, where the robot struggles to find safe motions and becomes stuck within the crowd. Prior research has recognized that modeling human-human and human-robot interactions is a critical aspect in addressing the FRP~\cite{mavrogiannis2023core, samavi2024sicnav}.

Early model-based reactive planners~\cite{helbing1995social, berg2011orca} attempted reciprocal collision avoidance or force-based models, but failed to capture intricate interactions in dense human environments, resulting in FRP for the robot. 
Recent reinforcement learning-based (RL) approaches~\cite{chen2017decentralized, chen2019sarl} learn socially compliant behaviors by capturing interactions implicitly from simulation data, but mapping crowd states to low-level actions can be difficult to regulate safety~\cite{ernst2009rlmpc}.

We mitigate this issue by leveraging \textit{crowd flow}. When moving in dense crowds, humans can make progress by following others moving in compatible directions~\cite{liao2025following} instead of being stalled by repeated individual avoidance maneuvers. This observation motivates our hypothesis that in dense human environments, following a moving crowd can be an effective navigation strategy, enabling robots to maintain consistent progress toward their goals in dense crowds.

\begin{figure}[t]
\centering
\includegraphics[width=.90\linewidth]{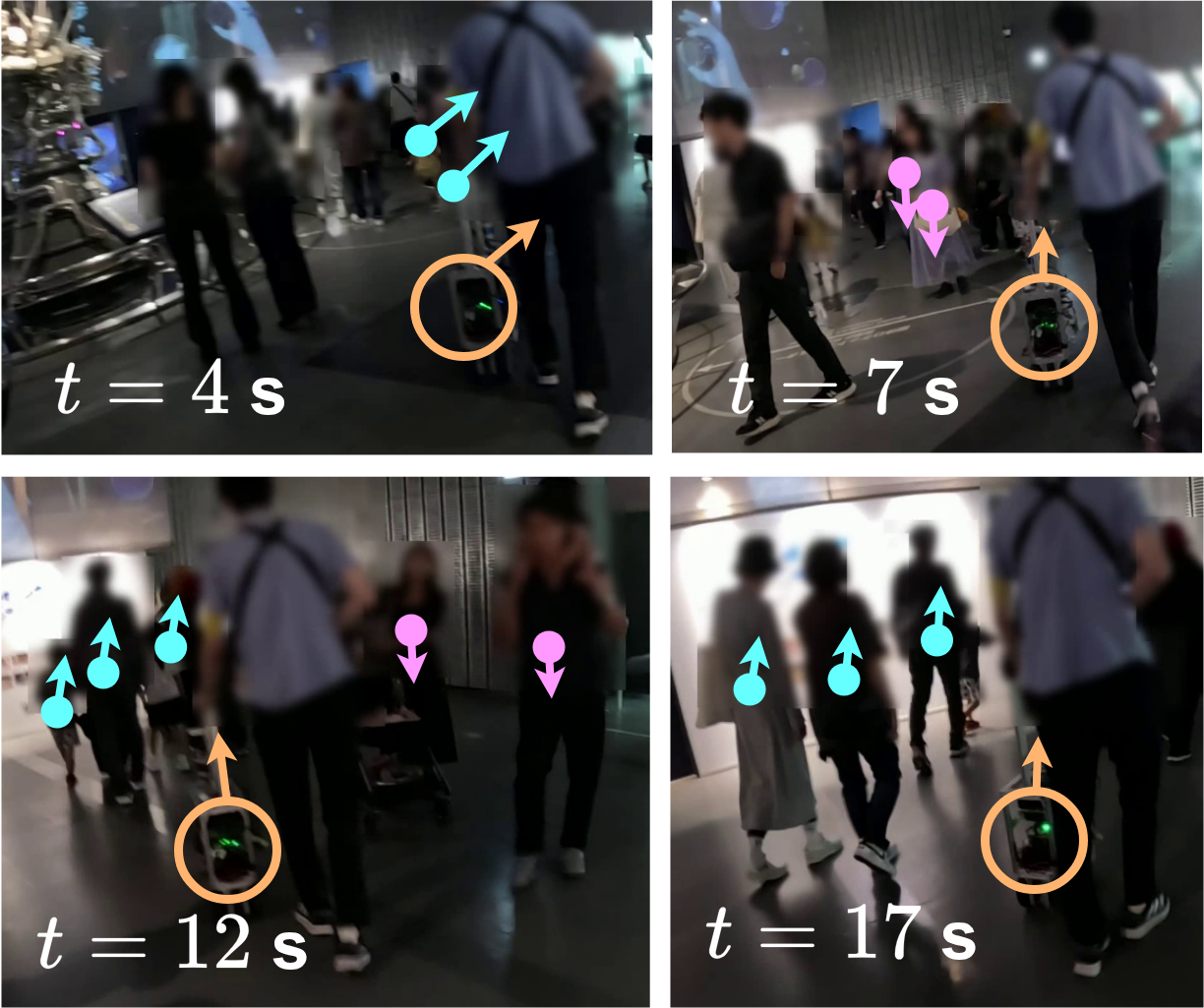}
\caption{Example of HiCrowd navigating in a real-world dense crowd environment. The mobile robot platform is marked with an \textbf{orange circle}. Pedestrians moving in a similar direction are shown in \textbf{cyan}, while those walking in the opposite direction are shown in \textbf{magenta}.
At $t=4$, the robot aligns with the pedestrian flow in front and follows it to the right, without yet observing the opposing group that appears later. When the oncoming pedestrians (magenta) appear at $t=7$, the robot, already aligned with the group ahead, smoothly follows the flow to the right and avoids collision.
By $t=12$ and $t=17$, the robot maintains alignment with the surrounding flow and avoids another group, achieving safe and efficient progress.
}
\label{fig:teaser}
\vspace{-1em}
\end{figure}

In this work, we propose HiCrowd, a hierarchical architecture that integrates reinforcement learning (RL) with model predictive control (MPC) to enable efficient and safe navigation in dense human environments, as shown in Fig.~\ref{fig:teaser}. HiCrowd's high-level RL policy predicts a socially aware \emph{follow point} that guides the robot to align with a suitable pedestrian group (i.e., the local crowd flow). The low-level MPC then optimizes a short-horizon trajectory to track this guidance while ensuring dynamic feasibility and safety with respect to nearby humans. This hierarchical design separates long-term decision making (group following and goal progress) from short-term motion control (collision avoidance), enabling consistent progress toward the goal in dense crowds while ensuring reliable safety in highly dynamic environments.

We evaluate HiCrowd against both classical and learning-based baselines, including the reactive planner ORCA~\cite{berg2011orca}, and the RL-based methods SARL~\cite{chen2019sarl} and CrowdAttn~\cite{liu2023attn}. We include an MPC-only ablation to isolate the contribution of the crowd-following module. 
Experiments are conducted in two settings: 
(1) an \emph{offline} setting, where humans replay recorded motions without reacting to the robot, and
(2) an \emph{online} setting, where humans (controlled via ORCA) react to the robot in simulation.
We use the ETH-UCY dataset~\cite{lerner2007crowds, pellegrini2010improving} and a synthetic crowd dataset to test navigation performance in dense human environments. 

Experimental results show that HiCrowd outperforms baseline methods, achieving higher success rates and more efficient navigation. It also maintains safe social distances and significantly reduces freezing in dense crowds. The qualitative analysis highlights that HiCrowd consistently aligns with crowd flows and reaches goals with minimal delay. In contrast, baseline methods tend to become stuck in dense opposing flows, leading to abrupt reactive actions and frequent freezing. Finally, we deployed HiCrowd on a mobile robot in two real-world crowded environments: a public museum and Expo 2025 Osaka. Without any retraining, the robot successfully navigated dense pedestrian flows, robustly demonstrating safe and socially aware behavior.

In summary, we make the following contributions:
\begin{itemize}
    \item Demonstrating that leveraging human motion as guidance, rather than solely as obstacles, provides a powerful principle for safe and efficient crowd navigation.
    \item Introducing a hierarchical RL-MPC framework where the RL agent is trained with a crowd-following reward that encourages alignment with compatible crowd flow.
    \item Showing that the crowd-following reward accelerates learning, improves performance, and can be effectively deployed in the real world.
\end{itemize}

%



\section{Related Work}
\label{sec:relatedwork}
Robot navigation in dynamic crowds has been studied for decades. 
Classical rule-based approaches such as 
social force~\cite{helbing1995social} and Optimal Reciprocal Collision Avoidance (ORCA)~\cite{berg2011orca} model interactions through attractive and repulsive forces or reciprocal velocity obstacles. While effective in structured scenarios, their performance degrades in complex crowd settings, especially when pedestrians behave unpredictably, resulting in the freezing robot problem~\cite{trautman2010unfreezing}. 

Instead of explicitly modeling agent behaviors, recent approaches have used reinforcement learning (RL) to implicitly capture complex inter-agent interactions.
Deep RL methods~\cite{chen2017decentralized} have been introduced for decentralized multi-agent collision avoidance, where each agent independently learns safe behaviors without explicit communication. Building on this line of work, SARL~\cite{chen2019sarl}, an attention-based Deep RL framework, has been proposed to enable robots to reason about interactions with surrounding pedestrians in a socially compliant manner.
More recently, \textcite{liu2023attn, liu2026height} extended this direction by using a spatio-temporal attention graph to encode interactions, while \textcite{xie2023drlvo} combined Deep RL with velocity obstacles to navigate crowded dynamic scenes. These works demonstrate the potential of Deep RL to learn socially aware navigation strategies directly from simulated data, though the ability to generalize to dense and diverse crowd environments remains an open challenge.

Beyond RL policies, hierarchical reinforcement learning (HRL) has been widely explored in robotics as a way to improve learning efficiency. \textcite{nachum2018data} propose a HRL method that generates subgoals to enable efficient robot control for navigation tasks. \textcite{nasiriany2022augmenting} replace the low-level agent in HRL with predefined parametric primitives to separate primitives from parameter estimation. \textcite{yang2024learning} apply HRL to combine a sequence of learned parameterized primitives for robotic manipulation. These works demonstrate how hierarchical structures can simplify decision making by decomposing long-horizon problems into subgoals or primitives, a principle that is also relevant for navigation in dynamic crowds.

Early work demonstrated the feasibility of integrating RL with MPC for control tasks~\cite{ernst2009rlmpc}. \textcite{wangba2020exploring} proposed using policy networks to guide sampling-based MPC by generating candidate actions for high-dimensional locomotion control tasks. 
\textcite{brito2021gompc} introduced an RL subgoal planner to guide an MPC local planner, rewarding goal-reaching while penalizing collisions. Most recently, \textcite{han2025drmpc} proposed DR-MPC, integrating MPC path tracking with model-free Deep RL to address the challenges of large data requirements and unsafe initial behaviors. 

Moreover, human flow is important for robots to conform to surrounding motion patterns~\cite{zhu2026nemomap}. Previous approaches have leveraged pedestrian group formations, a simple form of crowd flow. \textcite{wang2022group} and \textcite{katyal2022group} considered groups in their cost terms in MPC- and RL-based frameworks respectively. However, these methods optimized goal reaching while avoiding group intrusions, without explicitly considering the integration of underlying crowd flow.
\textcite{swaminathan2018down} incorporated learned site-specific motion patterns into an RRT\textsuperscript{*} planner, but these patterns are computed offline, making real-time adaptation difficult.~\cite{zhu2025fast}.

Recent work leverages online observations of pedestrians to guide navigation. 
Liao et al.~\cite{liao2025following} propose a people-as-planners framework that selects a human leader via rule-based evaluation.
But relying on a single leader and rule-based selection limits generalization to diverse real-world scenarios. 
In this work, we bridge this gap by focusing on \emph{crowd flow alignment} and learning a socially aware follow point with a high-level RL policy, enabling safe and efficient navigation in dense human environments.

\section{Method}
\label{sec:method}

\begin{figure}[t]
\centering
\includegraphics[clip,trim= 15mm 34mm 25mm 11mm,height=20mm]{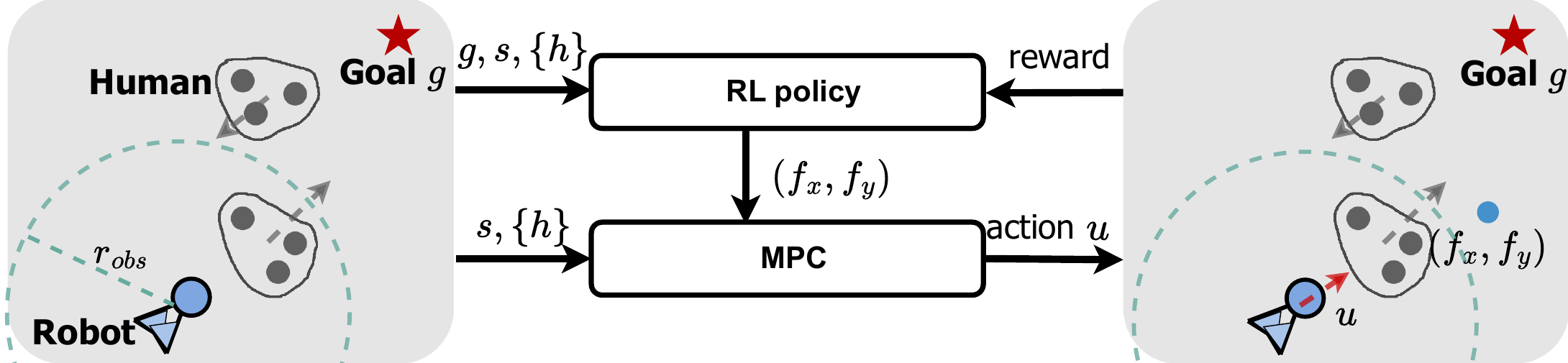}
\caption{Method overview. The robot observes nearby humans within its sensing radius $r_\mathrm{obs}$ along with its own state $\boldsymbol{s}$ and goal $\boldsymbol{g}$. A high-level RL policy outputs a follow point $(f_x,f_y)$ that guides the robot to align with a suitable crowd flow. The low-level MPC controller then generates a control action $\boldsymbol{u}$ to track this follow point while ensuring collision avoidance and dynamic feasibility. The RL policy is updated based on rewards that combine goal reaching, progress, and crowd following.}
\label{fig:method}
\vspace{-1em}
\end{figure}

We propose a hierarchical crowd navigation approach that integrates reinforcement learning (RL) with model predictive control (MPC) for robot navigation in dense human environments. An overview of the framework is shown in \cref{fig:method}. 
At the high level, the RL policy produces a \emph{follow point} to guide the robot to align with a suitable group in the current crowd. 
At the low level, the MPC controller uses this follow point as a reference while optimizing a short-horizon trajectory that ensures dynamic feasibility and safety with respect to nearby humans. 
The framework separates long-term decision making from short-term motion control, enabling the robot to progress consistently toward the goal even in dense crowds, mitigating the freezing robot problem.

\subsection{Problem Definition}
We consider the problem of robot navigation in dense human environments. The robot navigates in a 2D workspace $\mathcal{W} \subset \mathbb{R}^2$ among human agents.

At time $t$, the robot state is represented by its position and orientation, $\boldsymbol{s_t}=(p_{x,t}^r, p_{y,t}^r, \theta_t)$,
and the robot is tasked with reaching a goal $\boldsymbol{g}=(g_x, g_y)$.
The robot does not know the humans' goals, but can observe them within a sensing range $r_{\mathrm{obs}}$.
Within $r_{\mathrm{obs}}$, the robot observes $n$ humans, each represented by position and velocity, 
$\boldsymbol{h_t^i}=(p_{x,t}^i, p_{y,t}^i, v_{x,t}^i, v_{y,t}^i), \quad i=1, \dots, n$
and $\boldsymbol{H_t}=(\boldsymbol{h_t^1}, \boldsymbol{h_t^2},\dots,\boldsymbol{h_t^n})$ denotes the set of human states at time $t$. Humans may form groups when spatially close and moving in similar directions.

At each step, the robot executes a control input $\boldsymbol{u_t} = (v_t, \omega_t)$
where $v_t$ is the linear velocity and $\omega_t$ is the angular velocity. 
The robot state is updated according to a unicycle model with time step $\Delta t$: 
\begin{equation}
\begin{aligned}
x_{t+1} &= x_t + v_t \cos(\theta_t)\,\Delta t, \\
y_{t+1} &= y_t + v_t \sin(\theta_t)\,\Delta t, \\
\theta_{t+1} &= \theta_t + \omega_t \,\Delta t .
\end{aligned}
\label{eq:transition}
\end{equation}

For $\omega_t \neq 0$, the exact displacement can be expressed as an arc of radius $v_t / \omega_t$.

\subsection{Reinforcement Learning Module}
The reinforcement learning (RL) policy acts as a high-level decision maker, responsible for generating a \emph{follow point} that guides the robot through the crowd. Unlike RL-based motion planners that directly output control action~\cite{chen2019sarl, liu2023attn}, our RL agent predicts a point in robot-centric coordinates. This follow point serves as a target reference for the low-level MPC controller, which will be described later in \cref{III-C}. 

The RL policy observes the robot state $\boldsymbol{s}_t$, the robot goal $\boldsymbol{g}$, 
and the set of nearby human states $\boldsymbol{H}_t$ within the observation range $r_{\mathrm{obs}}$. The number of input human states is fixed by zero-padding the human state array to a predefined maximum, maintaining a constant input dimension. The action of the RL policy is a 2D follow point 
$\boldsymbol{f}_t = (f_{x,t}, f_{y,t})$ in the robot's local frame. 
Formally as, $\boldsymbol{f}_t \sim \pi(\cdot \mid \boldsymbol{s_t}, \boldsymbol{H_t}, \boldsymbol{g})$.
The optimal policy $\pi^*$ maximizes the expected return: 
$
    \pi^* = \arg\max_{\pi} \;
    \mathbb{E}_{\pi}\!\left[
        \sum_{k=0}^{\infty} \gamma^k r_{t+k}
    \right]
$
, where $r_{t+k}$ is the reward at time $t+k$, and $\gamma \in [0,1)$ is the discount factor.
At each time step $t$, the reward is defined as the weighted sum of three components: 
\begin{equation}
    r_t = r_t^{\mathrm{goal}} + \lambda_{d} r_t^{\mathrm{dist}} + \lambda_{f} r_t^{\mathrm{follow}},
\end{equation}

\noindent\textbf{Goal reaching sparse reward} ($r_t^{\mathrm{goal}}$). The agent receives positive reward when it reaches goal location $\boldsymbol{g}$.

\noindent\textbf{Goal reaching dense reward} ($r_t^{\mathrm{dist}}$). 
The agent is encouraged to be closer to the goal, defined as
$r_t^{\mathrm{dist}} = \|\boldsymbol{p}_{t-1} - \boldsymbol{g}\| - \|\boldsymbol{p}_t - \boldsymbol{g}\|$, where $\boldsymbol{p}_t = (p_{x,t}^r, p_{y,t}^r)$, i.e. the robot position at time $t$. This term is weighted by $\lambda_d$.

\noindent\textbf{Crowd following reward} ($r_t^\mathrm{follow}$). 
To compute $r_t^\mathrm{follow}$, we first cluster the $n$ observed humans into groups based on spatial proximity and motion similarity. 
We apply the DBSCAN algorithm~\cite{ester1996dbscan} to human states $\{\boldsymbol{h}_t^i\}_{i=1}^n$ with thresholds $(\varepsilon_p, \varepsilon_\theta, \varepsilon_v)$ on position, orientation, and speed, respectively.
DBSCAN assigns each human $i$ a group label $\ell_t^i$. 
Each group is then defined as $G_j = \{ i \mid \ell_t^i = j\}$, 
where a group may include one or more people. 
For each group $G_j$, we compute scores that measure how well the 
robot's current motion matches the group's average motion:
$s^{\mathrm{speed}}_j = \exp\!\left(-\big(\tfrac{\Delta v_j}{\sigma_v}\big)^2\right)$, 
$s^{\mathrm{angle}}_j = \exp\!\left(-\big(\tfrac{\Delta \theta_j}{\sigma_\theta}\big)^2\right)$, 
$s^{\mathrm{dist}}_j = \exp\!\left(- d_j^2 \right)$,
where $\Delta v_j$ and $\Delta \theta_j$ are the differences between the 
robot's speed and heading and the average speed and heading of group $G_j$, 
and $d_j$ is the minimum distance from the robot to any member of group $G_j$.
The final crowd following reward is the maximum score across all groups:
\begin{equation}
    r_t^{\mathrm{follow}} = \max_j \left( s^{\mathrm{speed}}_j \cdot s^{\mathrm{angle}}_j \cdot s^{\mathrm{dist}}_j \right).
\end{equation}
Using the maximum score across groups, the robot aligns with the most suitable group, typically the one moving in a similar direction and toward a compatible destination, while retaining flexibility to switch groups as conditions change. This term is weighted by $\lambda_f$.

During training, the RL agent generates a follow point, which is passed to the MPC-based low-level controller. The MPC computes a safe robot control input $\boldsymbol{u}_t$ to execute, and the environment transitions to a new state as in~\cref{eq:transition}. The reward is computed based on the outcome of the MPC, and the RL policy is updated accordingly. This hierarchical design separates strategic decision-making (selecting a suitable group to follow) from low-level motion control (ensuring dynamic feasibility and collision avoidance).

\subsection{Model Predictive Control Module} \label{III-C}
At time $t$, the MPC module takes as input the robot state $\boldsymbol{s}_t$, 
the human states $\boldsymbol{H}_t$, and the follow point $\boldsymbol{f}_t$. 
It then optimizes a short-horizon trajectory by minimizing a cost function that balances 
progress toward the RL-provided follow point and maintaining safety with respect to nearby humans. The total cost is given by $J = J_{\mathrm{follow}} + J_{\mathrm{safe}}$.

\noindent\textbf{Follow cost} ($J_{\mathrm{follow}}$).
The follow cost encourages the robot to move towards the follow point, defined as the minimum Euclidean distance between the predicted trajectory 
and the RL-provided follow point $\boldsymbol{f}_t$:
\begin{equation}
    J_{\mathrm{follow}} = \min_{\tau \in [0, H-1]} \; \|\boldsymbol{p}^r_{t+\tau} - \boldsymbol{f}_t\|,
\end{equation}
where $\boldsymbol{p}^r_{t+\tau}$ is the predicted robot position at step $t+\tau$ 
over a planning horizon $H$. 

\noindent\textbf{Safety cost} ($J_{\mathrm{safe}}$).
The safety cost penalizes proximity to humans over the planning horizon. 
For each human $i$, we predict the future position using a constant velocity model (CVM), and compute the distance $d^i_{t+\tau}$ between the predicted robot position 
and the predicted human position at horizon step $\tau$. The safety cost is defined as 
\begin{equation}
    J_{\mathrm{safe}} = \sum_{\tau=0}^{H-1} \beta^\tau \sum_{i=1}^n 
    \exp\!\left( d_c - d^i_{t+\tau} \right),
\end{equation}
where $d_c$ is a collision radius, $\beta \in (0,1)$ is a discount factor that decrease the weights of future collision risks.

The MPC selects the control sequence $\boldsymbol{u}_{t:t+H-1}$ that minimizes $J$, 
and applies the first control input $\boldsymbol{u}_t$ to update the robot dynamics as in \cref{eq:transition}.

\section{Experiments}
\label{sec:experiments}
\vspace{-2mm}
\subsection{Datasets}
The evaluation is carried out in real-world datasets and a synthetic dataset.
\subsubsection*{ETH-UCY} The ETH-UCY dataset~\cite{lerner2007crowds, pellegrini2010improving} contains 5 subsets: ETH, HOTEL, UNIV, ZARA1 and ZARA2, with various motion patterns.
Similar to~\cite{wang2022group}, from these recordings, we construct navigation episodes by segmenting the trajectories into blocks that involve dense human interactions. A segment contains at least five pedestrians present in the scene. Across all scenarios, the number of pedestrians ranges from 5 to 22, with an average of 7.3 pedestrians per scene.

\subsubsection*{Synthetic} 
To evaluate navigation performance under even denser conditions, we construct a synthetic dataset simulating pedestrian flows. Each episode consists of two opposing flows of pedestrians. Groups are spawned at randomized intervals around \SI{5}{\second} from opposite sides of the scene, each containing 2 to 5 pedestrians. Within a group, each pedestrian is assigned an initial velocity sampled around a mean speed of \SI{0.9}{\meter\per\second}, with a random noise of $\pm0.05$\SI{}{\meter\per\second}, and moves with constant velocity.

We construct navigation episodes by applying the same segmentation procedure as used for ETH-UCY. Across all episodes, the number of pedestrians ranges from 6 to 63, with an average of 19.8 pedestrians per scene.

For both datasets, we generate multiple episodes per segment by uniformly sampling robot start and goal positions from start and goal regions placed on opposite sides of the dominant pedestrian flow to encourage interactions. We sample 5{,}000 training episodes from ETH-UCY and 3{,}000 from the synthetic dataset, in total across all segments. For evaluation, we generate 500 test episodes per dataset, and use the same test episodes for evaluating all methods.

\vspace{-1mm}
\subsection{Experimental Setup}
We evaluate in two settings that differ in whether pedestrians react to the robot.

\subsubsection*{Offline} Pedestrians replay recorded trajectories in the dataset and do not react to the robot. This setting evaluates whether a crowd navigation planner can reach the goal without causing deviations in pedestrian behavior. It also reflects real deployments with small robots that nearby pedestrians may overlook and fail to react in time.

\subsubsection*{Online} Pedestrian are initialized with the same start and goal locations as in the corresponding dataset segment. Their subsequent motion is simulated by ORCA~\cite{berg2011orca}, and pedestrians react to the robot and to each other, with maximum speed set to \SI{1.75}{\meter\per\second}.

\subsection{Baselines}
We compare with the following baseline methods.

\subsubsection*{MPC} The MPC baseline uses the same setup as the MPC component (\cref{III-C}) of HiCrowd, but instead of taking an RL-provided follow point as the reference, 
it directly uses the robot's goal location. This baseline evaluates the effect of the RL-guided crowd-following module in HiCrowd.

\subsubsection*{ORCA} Optimal Reciprocal Collision Avoidance (ORCA)~\cite{berg2011orca} serves as a reactive-based baseline, generating collision-free movements through reciprocal collision avoidance.
ORCA is evaluated as a baseline in both offline and online settings, and in online settings it also serves as the controller for pedestrians.

\subsubsection*{SARL} The Socially Attentive Reinforcement Learning (SARL)~\cite{chen2019sarl} learns a policy that maps crowd states directly to robot actions. This baseline evaluates the performance of the hierarchical RL-MPC design and the crowd-following strategy over a flat RL approach.

\subsubsection*{CrowdAttn} Crowd Navigation with Attention-Based Interaction Graph method~\cite{liu2023attn} augments RL-based navigation with a spatio-temporal attention graph. CrowdAttn serves as a recent state-of-the-art baseline. 
\vspace{-1mm}
\subsection{Evaluation Metrics}
The evaluation metrics include the success rate (\textbf{SR}), with failures arising from either collisions (\textbf{CR}) or timeouts (\textbf{TR}). In the case of a collision, the episode is immediately terminated as a failure. A timeout occurs when the robot fails to reach the goal within a time limit set to three times the duration of reaching the goal in a straight line in the dataset.

We also report the average navigation time in seconds (\textbf{NT}), the average path length for non-collision episodes in meters (\textbf{PL}), and the average minimum pedestrian distance in meters (\textbf{MP}). In addition, we measure the freezing frequency (\textbf{FF}), defined as the percentage of time steps in which the robot's velocity reaches zero (following~\cite{samavi2024sicnav}), indicating how often the robot becomes stuck among human agents.
\vspace{-1mm}
\subsection{Implementation Details}
The RL agent is trained for $100$K transitions using the Soft Actor--Critic (SAC) algorithm~\cite{haarnoja2018soft}. We use a discount factor of $\gamma = 0.99$ and a batch size of 256. The actor and critic are trained with learning rates of $3 \times 10^{-4}$ and $1 \times 10^{-3}$, respectively. The entropy coefficient is trained with a learning rate of $1 \times 10^{-3}$. We use Adam for all optimizations. Observation radius $r_\mathrm{obs}$ is set to \SI{5}{\meter}. The MPC module uses a planning horizon $H=10$ and discount factor $\beta=0.9$. The collision radius $d_c$ is \SI{0.5}{\meter} and is used as the threshold for collision detection. The ratio between high-level RL action and low-level MPC action is 10. For DBSCAN grouping, $\varepsilon_\theta=\pi/6$, $\varepsilon_v=1.0$, $\varepsilon_p=2$. For computing crowd following reward $r_t^\mathrm{follow}$, $\sigma_v=0.5$ and $\sigma_\theta=\pi/12$. Both goal reaching dense reward weight $\lambda_d$ and crowd following reward weight $\lambda_f$ are set to 1. In the ablation, we run experiments with $\lambda_f \in \{0, 1, 5\}$ to evaluate its effect on learning efficiency. The robot's maximum linear speed is \SI{1}{\meter\per\second} and angular speed is \SI{3.14}{\radian\per\second}.

For SARL training, the imitation learning part follows the same configurations as in~\cite{chen2019sarl}. The RL part is trained on both datasets in online and offline setup, for $5\times10^{5}$ transitions with a learning rate of 0.001. 
CrowdAttn is used with its pretrained model and the Gumbel Social Transformer (GST) for motion prediction, trained on ETH-UCY. For CrowdAttn and ORCA, we map holonomic action outputs to differential drive commands. The linear velocity is the velocity vector norm, and the angular velocity is the turning angle between the robot orientation and the holonomic action. This ensures comparability with our method and other baselines \footnote{We also trained CrowdAttn under a differential drive setting, but the results were suboptimal compared to those with the converted provided pretrained holonomic model, so we present pretrained model results.}. 




\vspace{-2mm}
\section{Results} \label{section-results}


\begin{figure*}[t]
\centering
\includegraphics[clip,trim= 0mm 12mm 0mm 10mm,height=10mm]{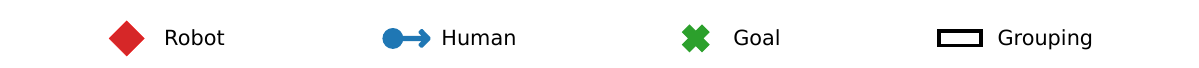}\\
\tileTop{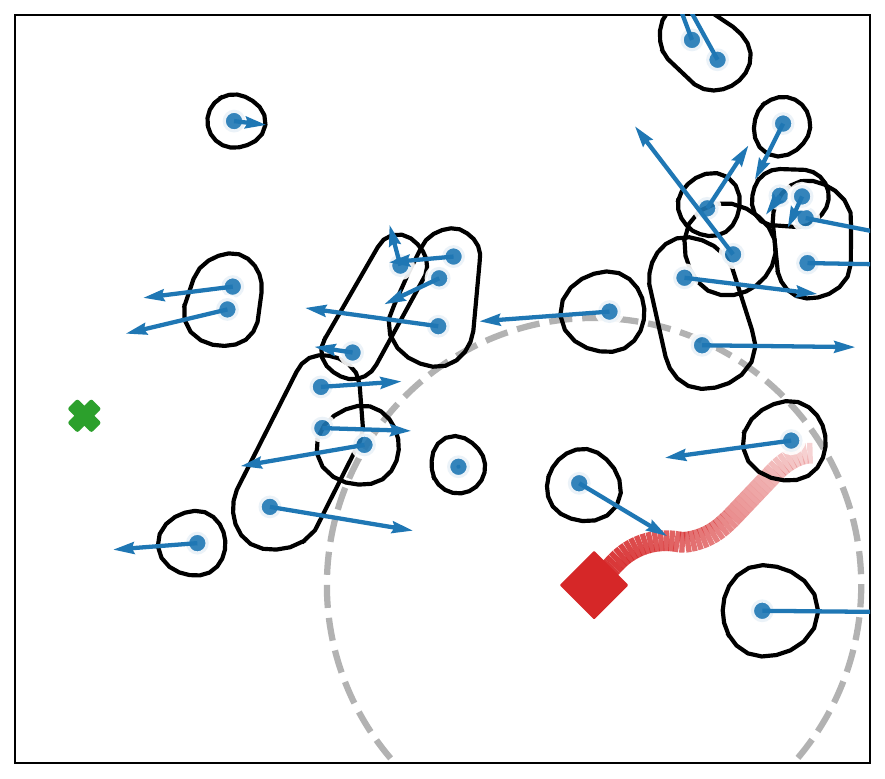}{HiCrowd (Ours)}{t=5}%
\tileTop{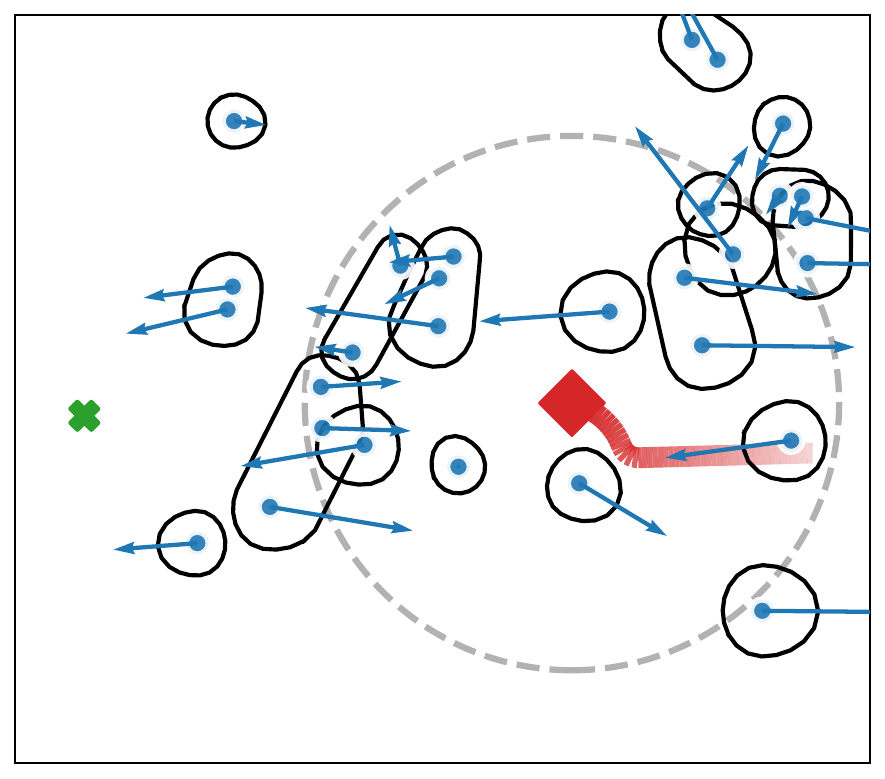}{MPC}{t=5}%
\tileTop{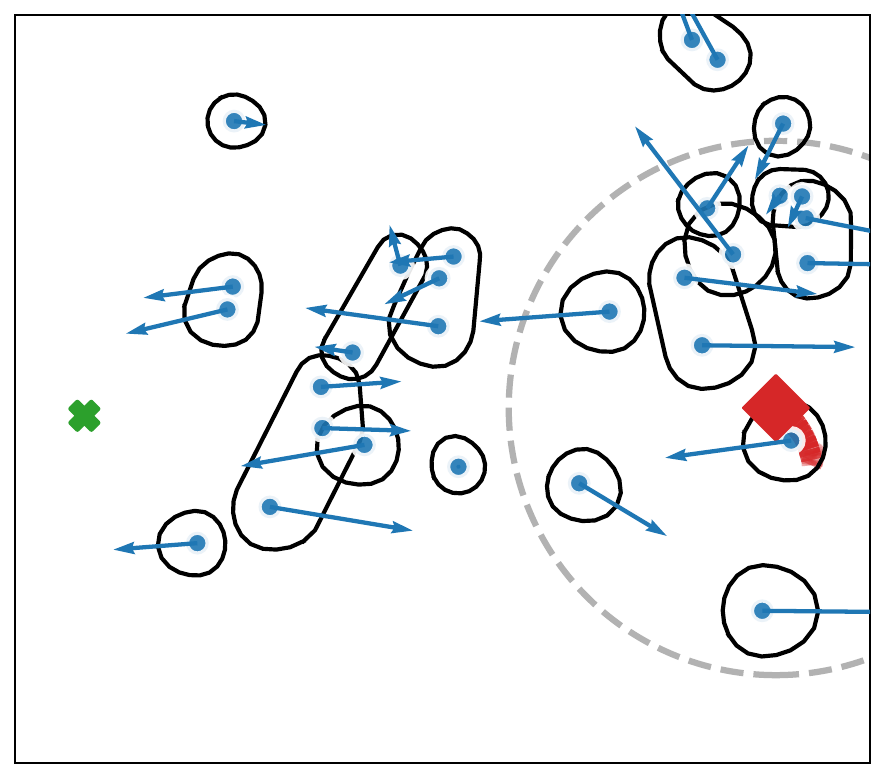}{ORCA}{t=5}%
\tileTop{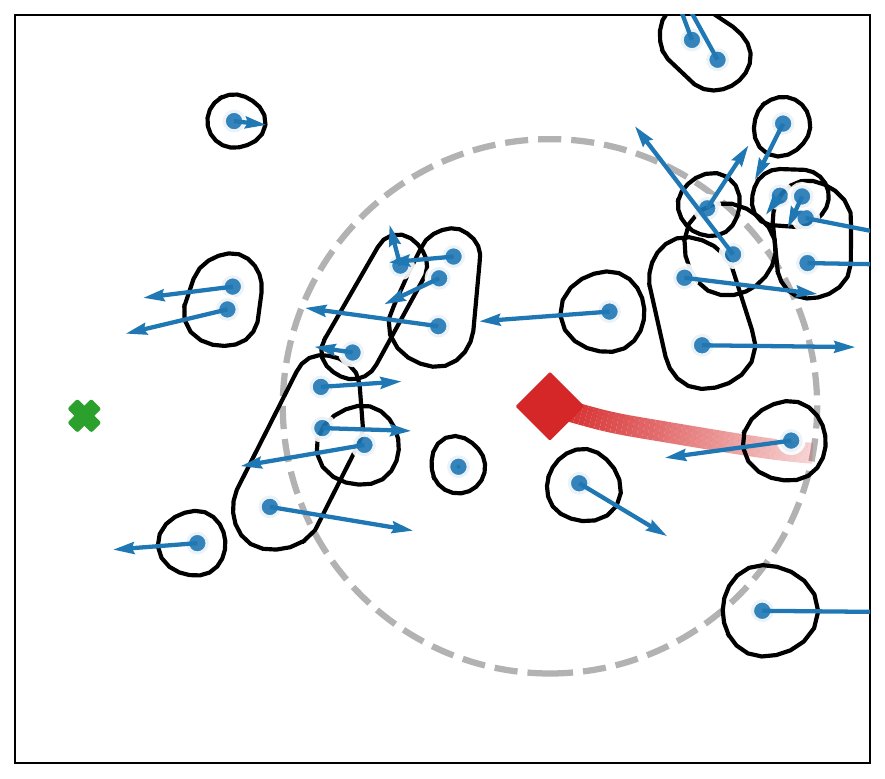}{CrowdAttn}{t=5}%
\tileTop{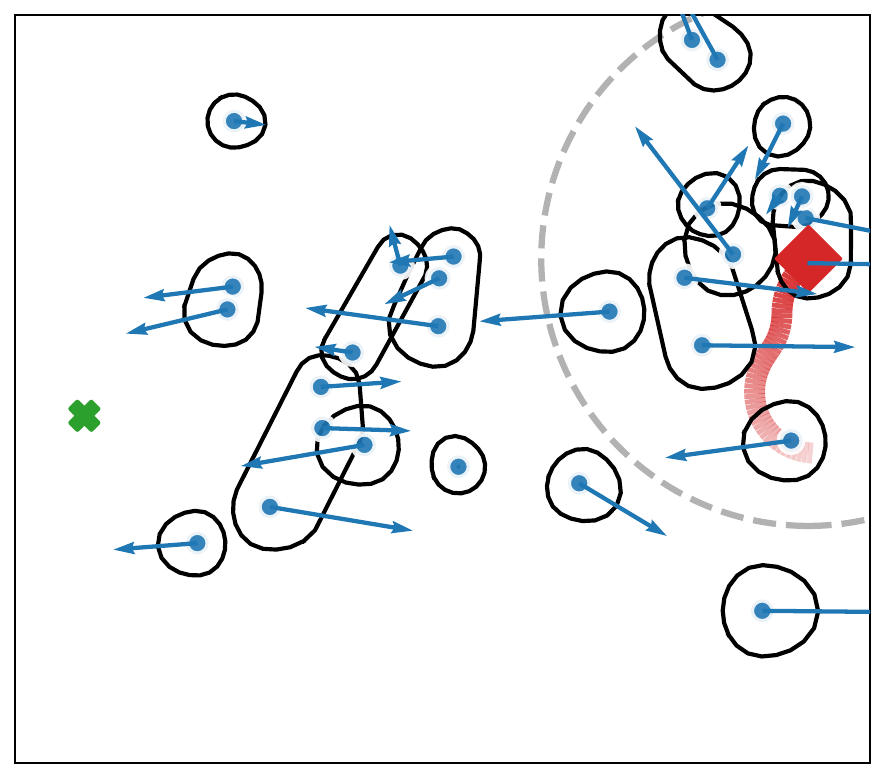}{SARL}{t=5}\\

\tile{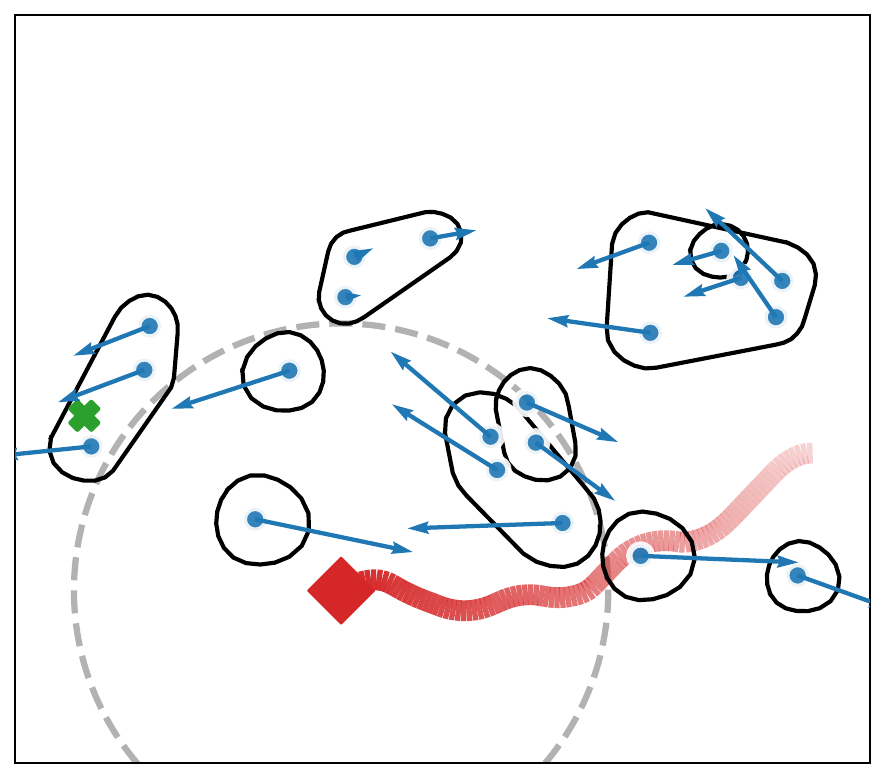}{t=10}%
\tile{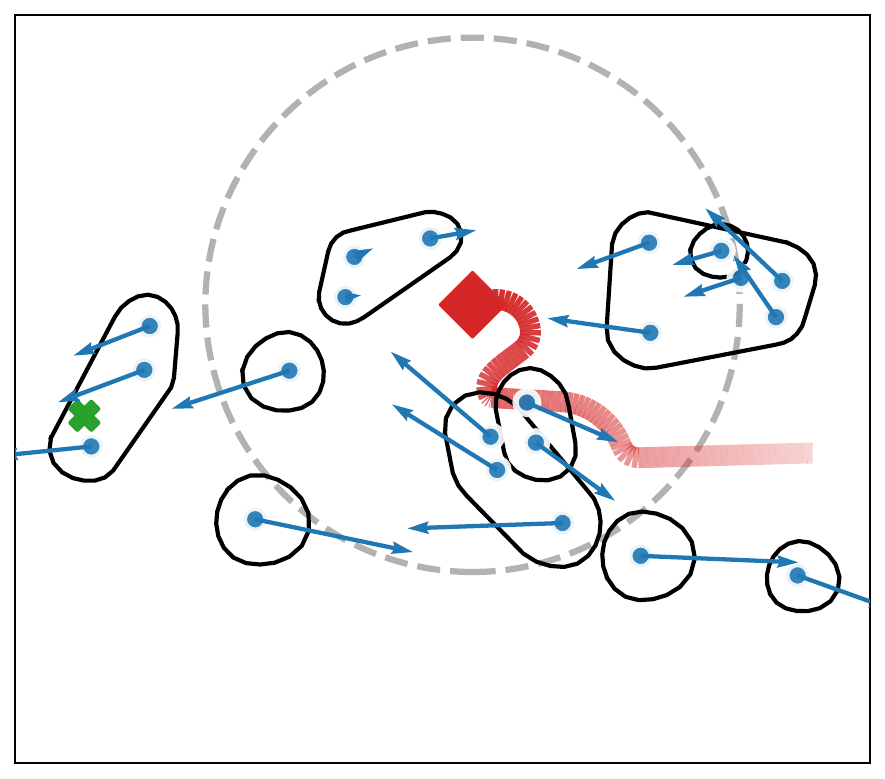}{t=10}%
\tile{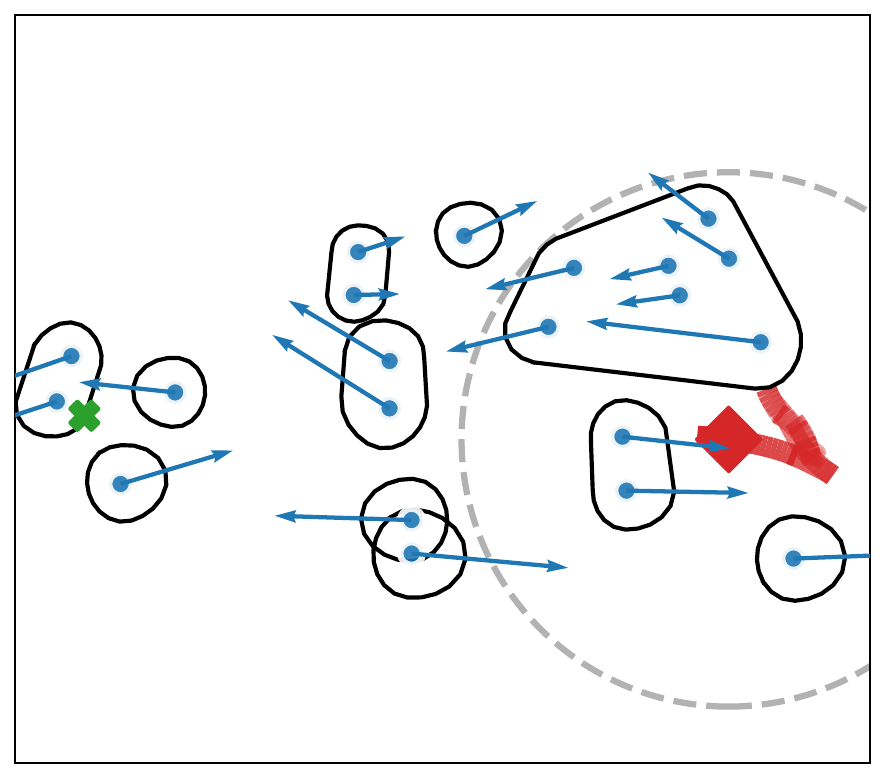}{t=12}%
\tile{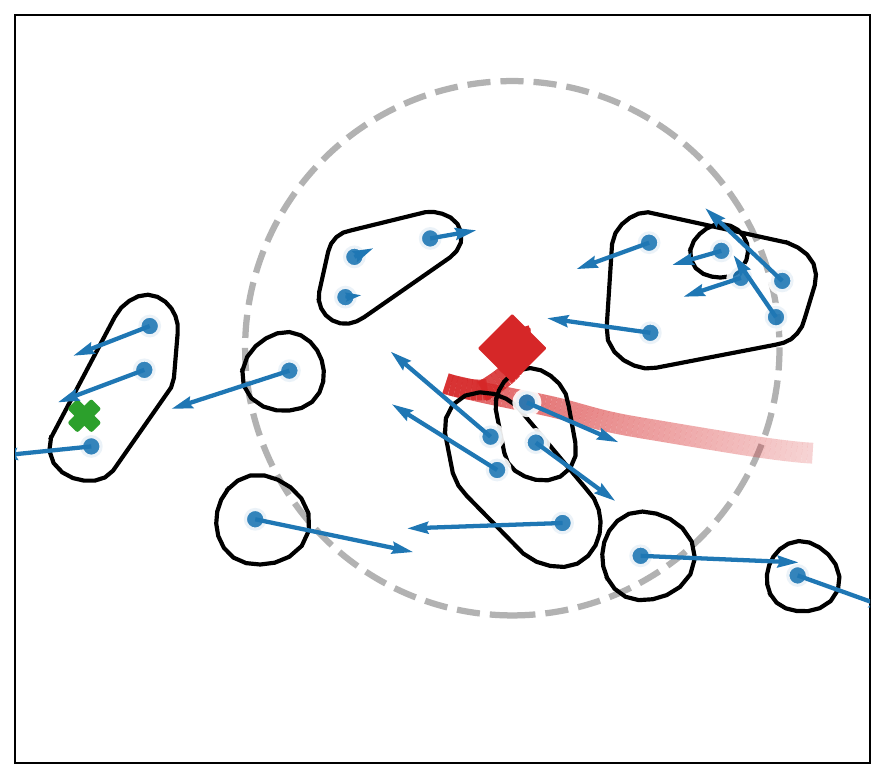}{t=10}%
\tile{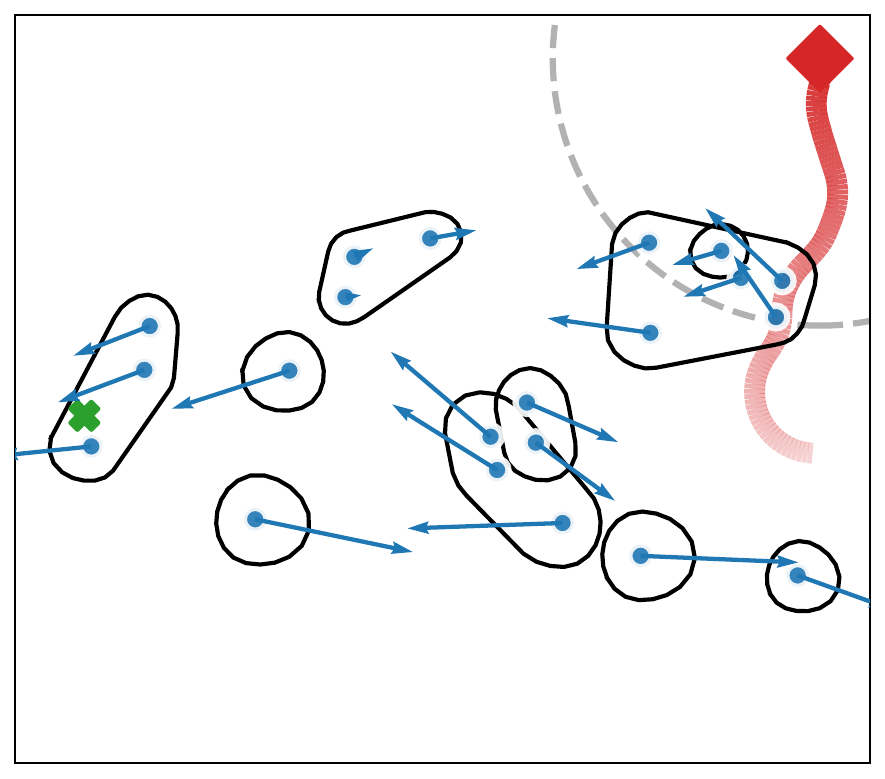}{t=10}\\

\tile{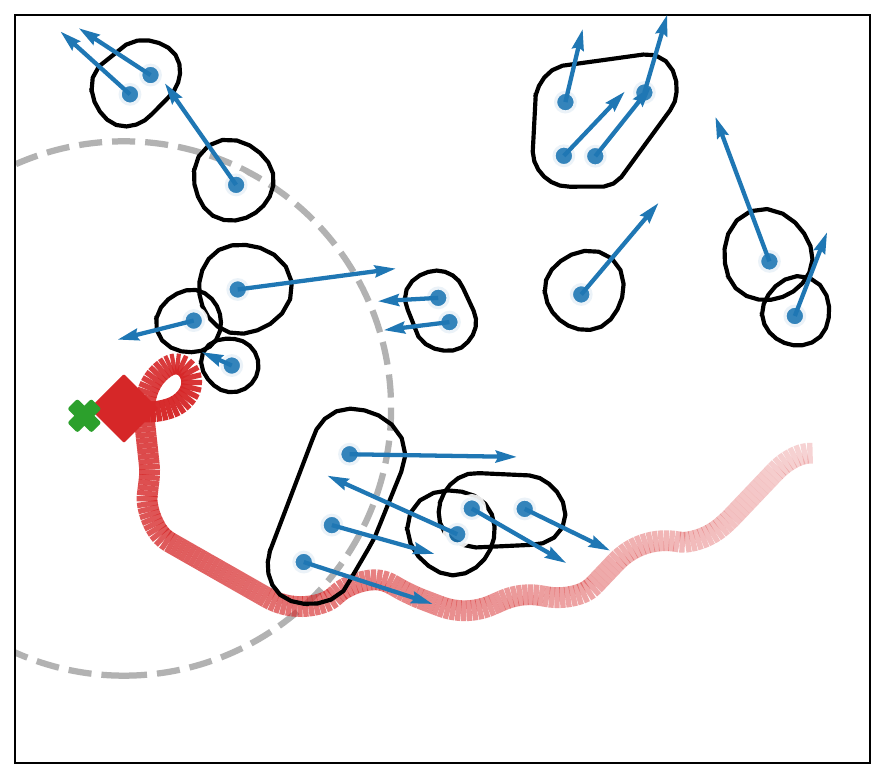}{t=20}%
\tile{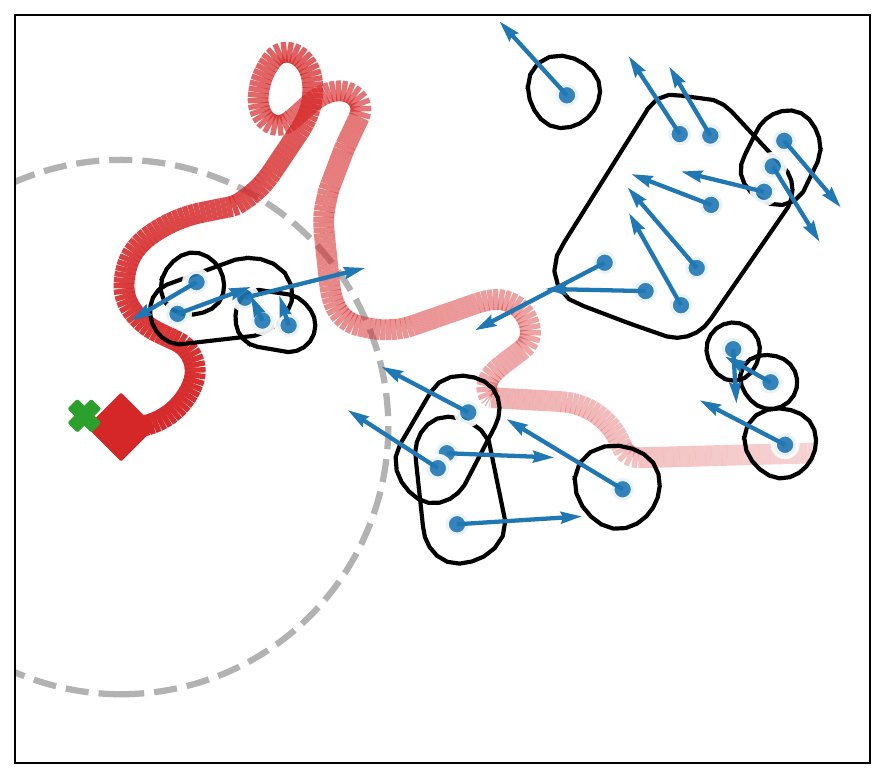}{t=31}%
\tile{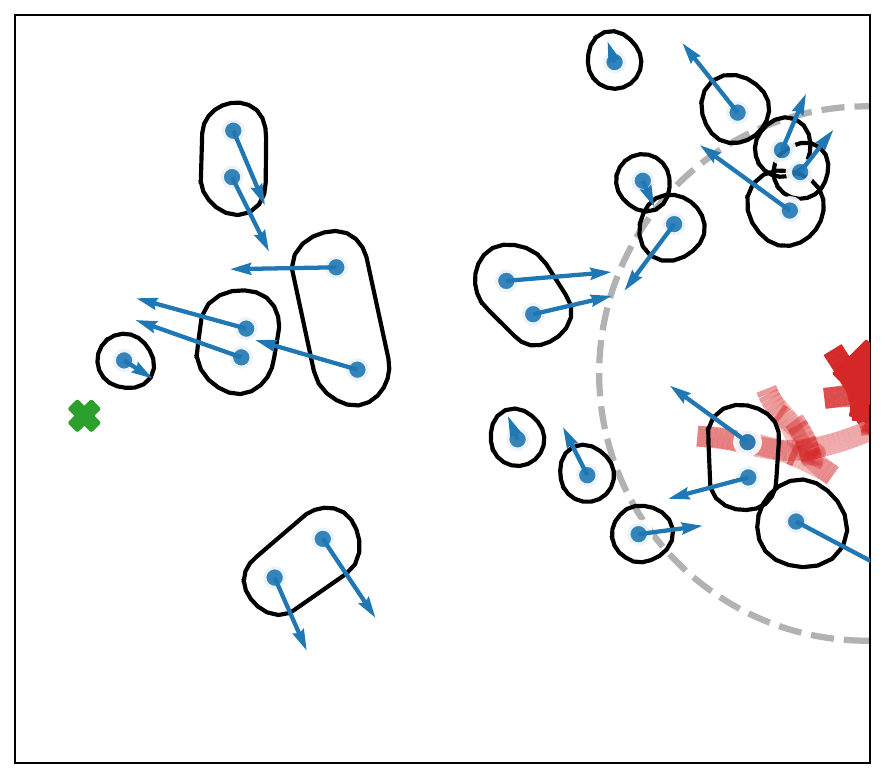}{t=60}%
\tile{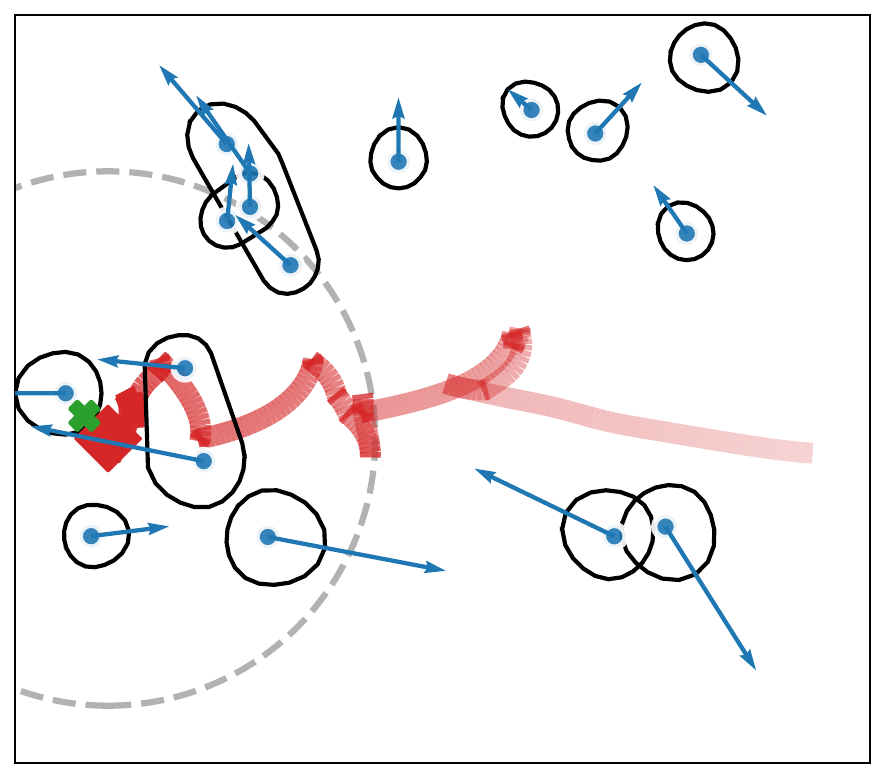}{t=39}%
\tile{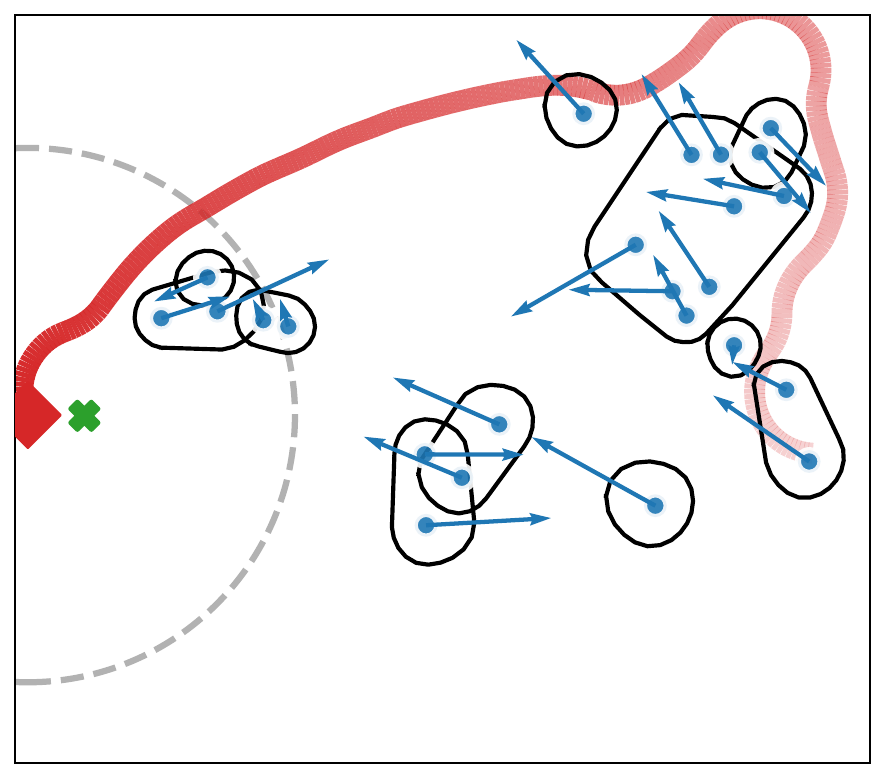}{t=31}\\

\caption{Example episode from the ETH-UCY dataset in the \textbf{offline} setting ($t$ in seconds). This case involves a dense crowd with both moving and static groups. Groups are visualized as convex hulls. Robot observation range is marked as a grey circle. Baseline methods take longer paths and more time to reach the goal, with multiple reactive avoidance actions. ORCA gets stuck near dense crowds. CrowdAttn freezes several times. And ORCA, SARL and CrowdAttn result in collisions. In contrast, HiCrowd successfully reaches the goal with a shorter navigation time.}
\label{fig:example_1904}
\end{figure*}

\begin{figure*}[th]
\centering

\tileTop{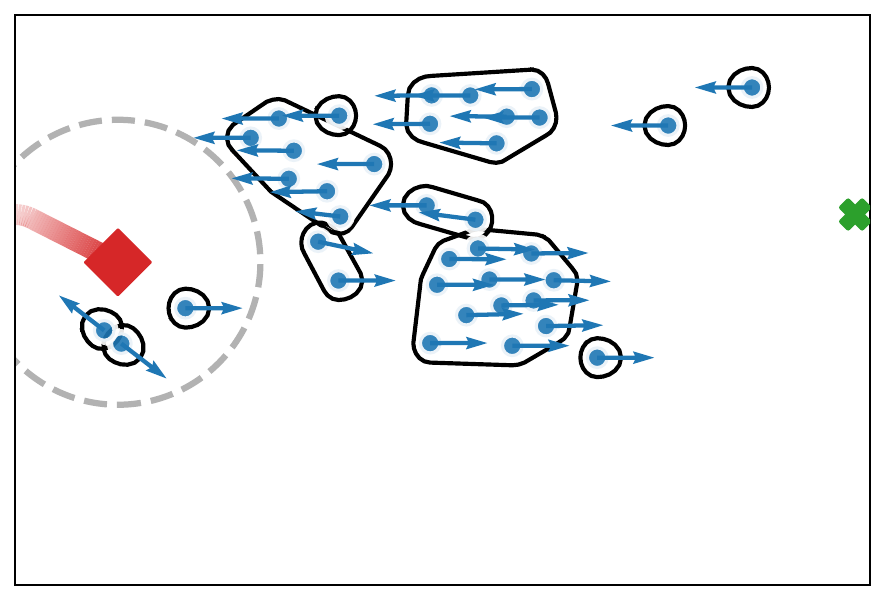}{HiCrowd (Ours)}{t=4}%
\tileTop{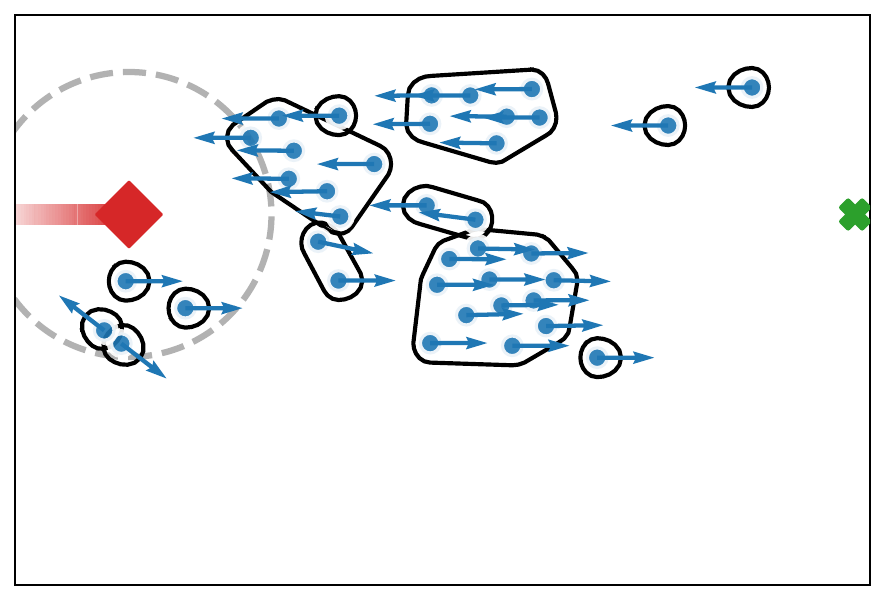}{MPC}{t=4}%
\tileTop{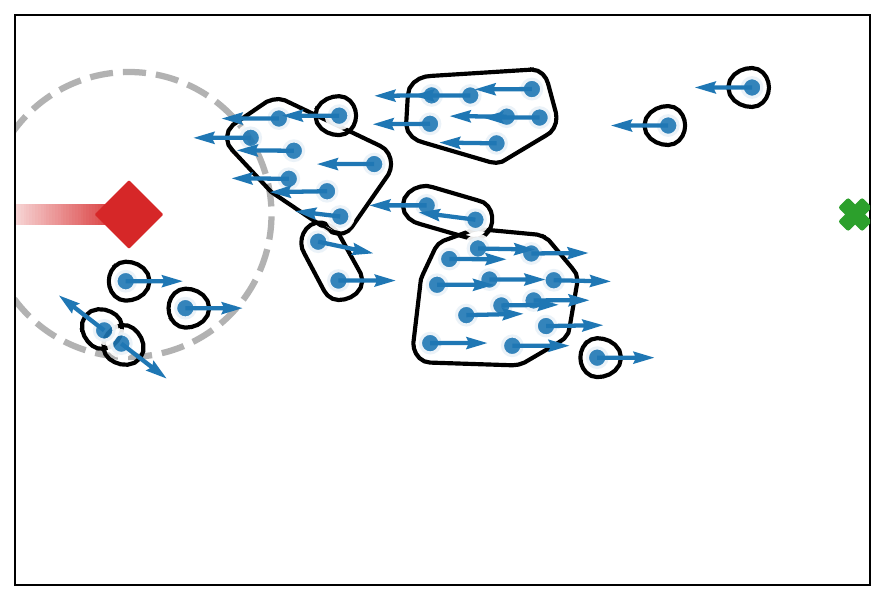}{ORCA}{t=4}%
\tileTop{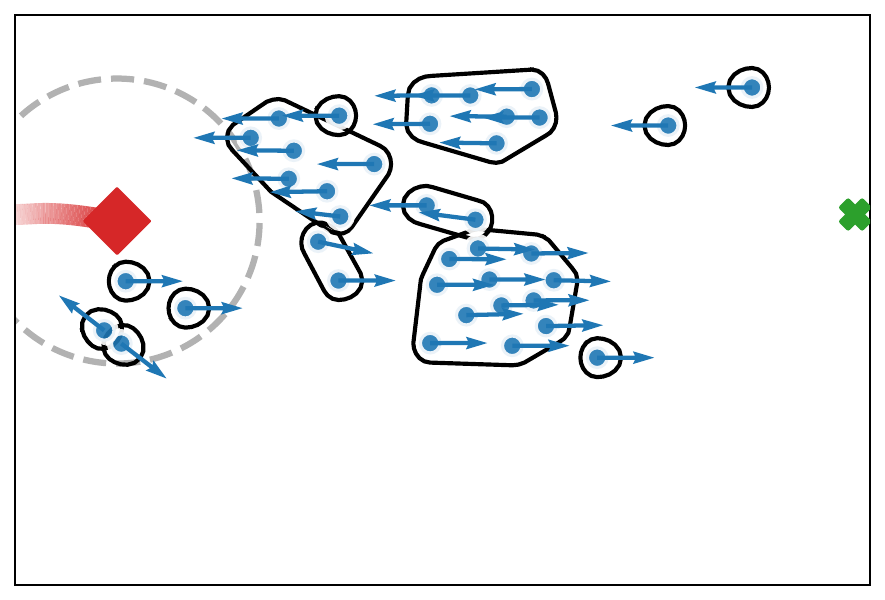}{CrowdAttn}{t=4}%
\tileTop{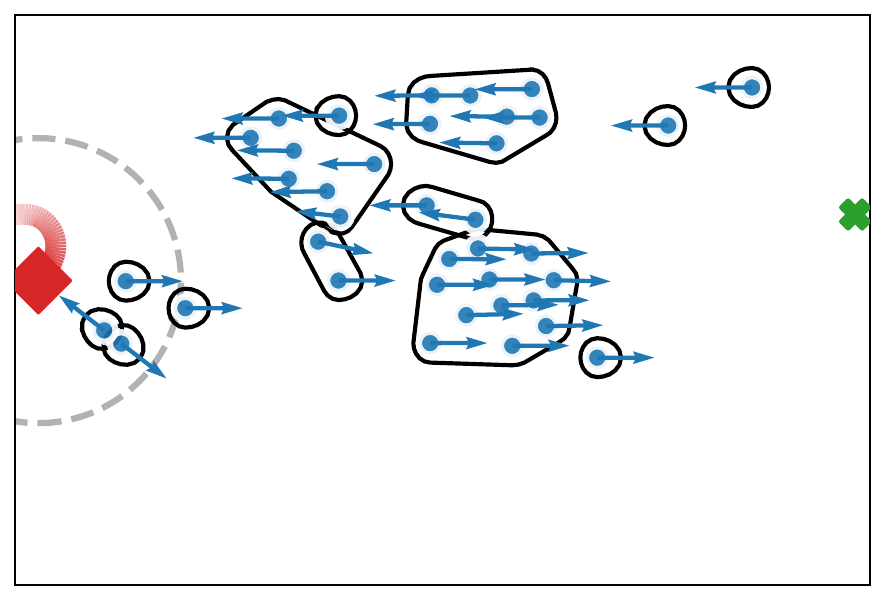}{SARL}{t=4}\\

\tile{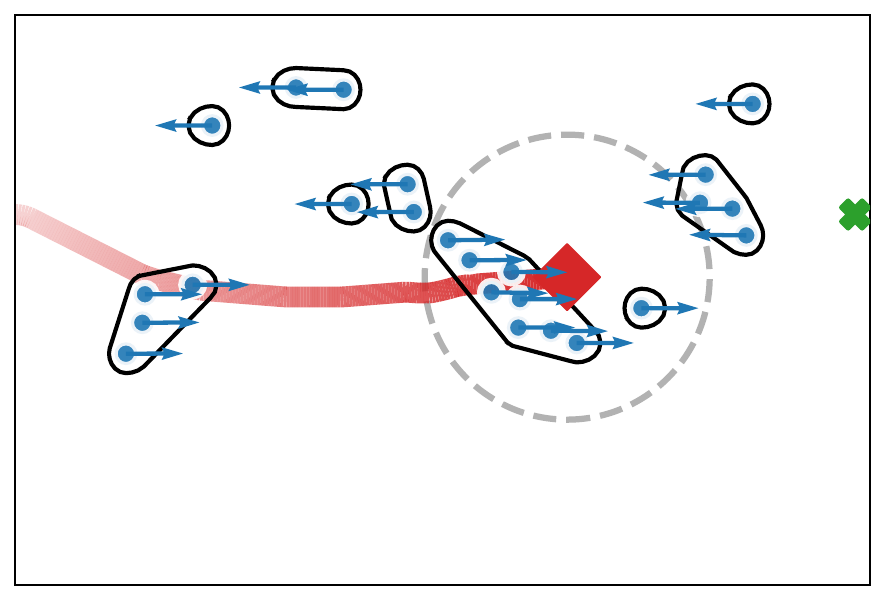}{t=20}%
\tile{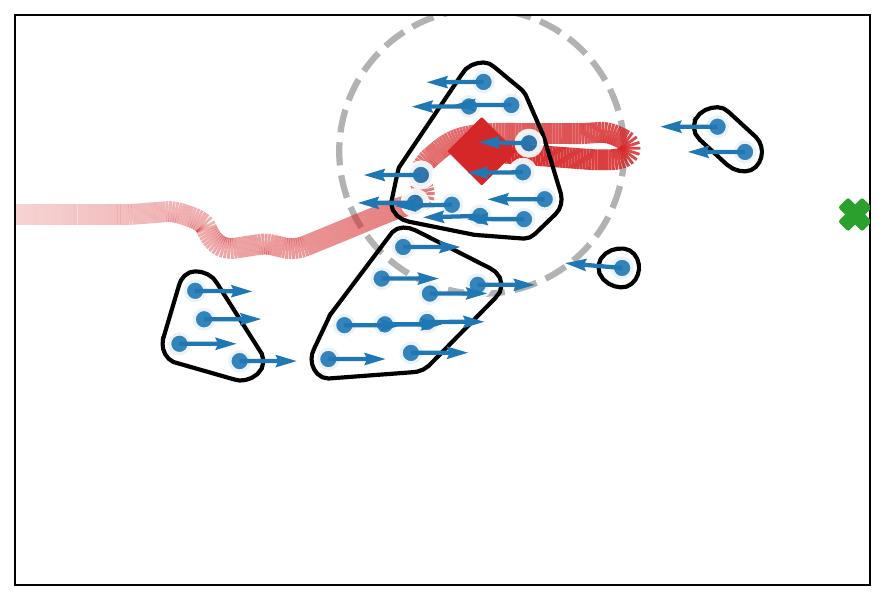}{t=30}%
\tile{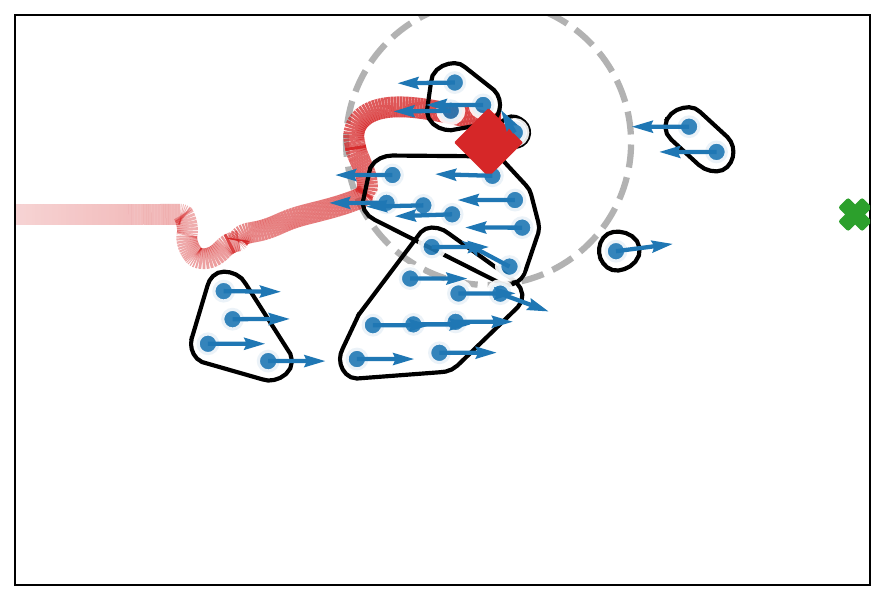}{t=31}%
\tile{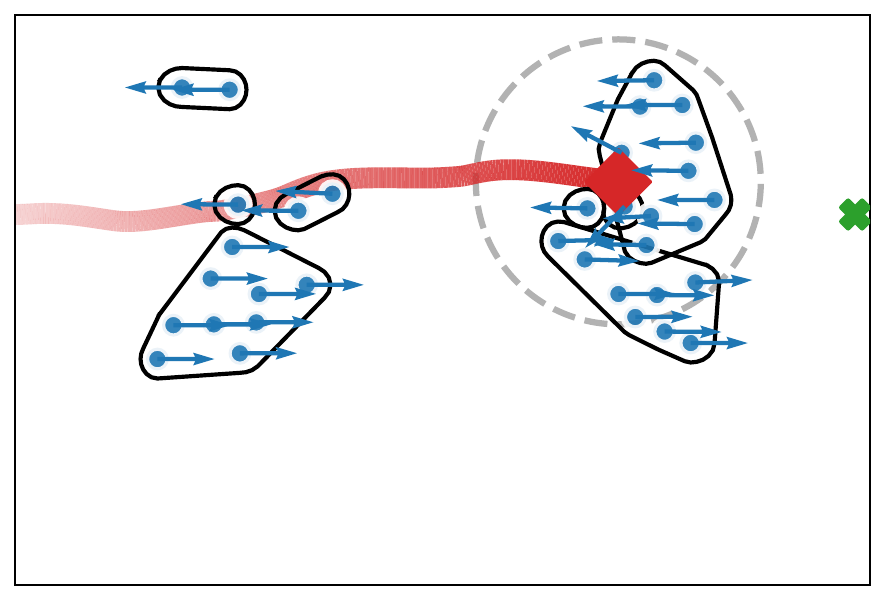}{t=24}%
\tile{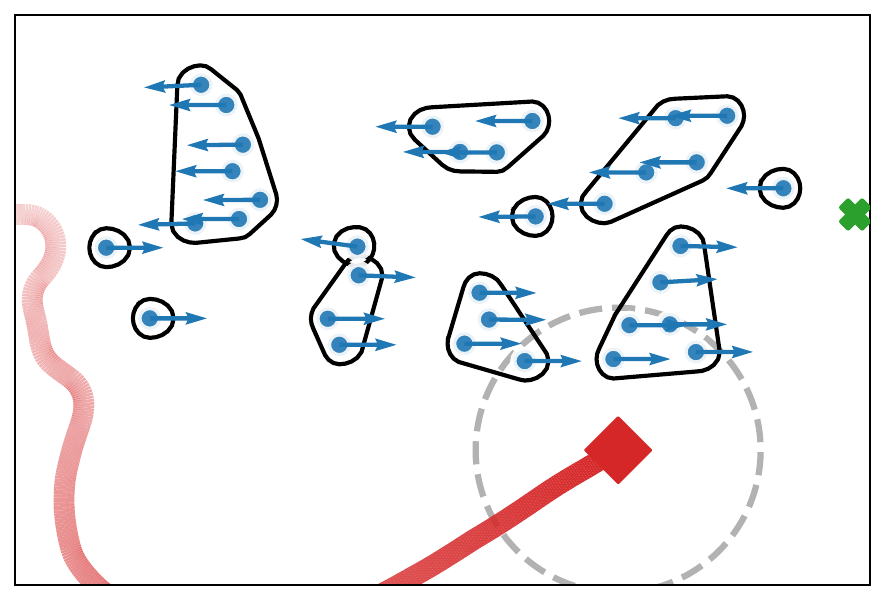}{t=40}\\

\tile{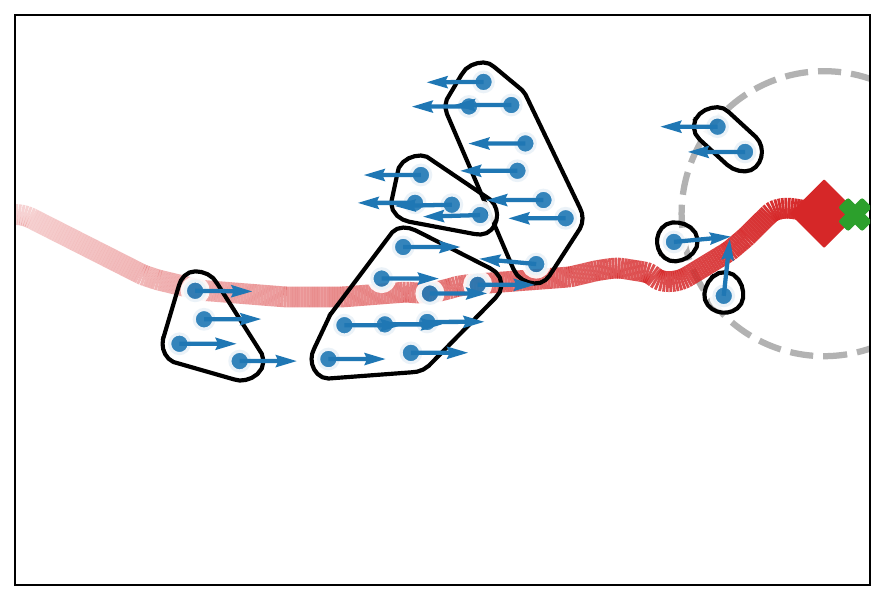}{t=30}%
\tile{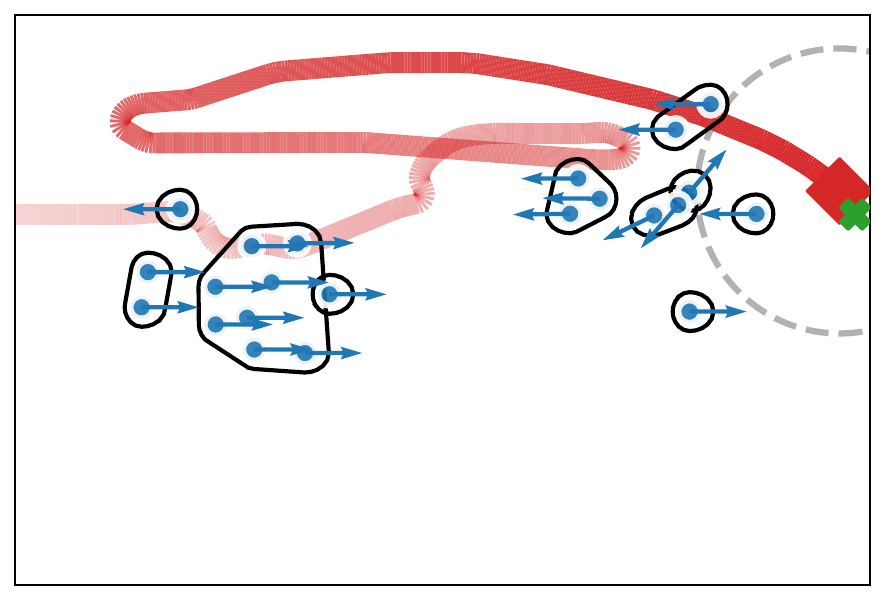}{t=70}%
\tile{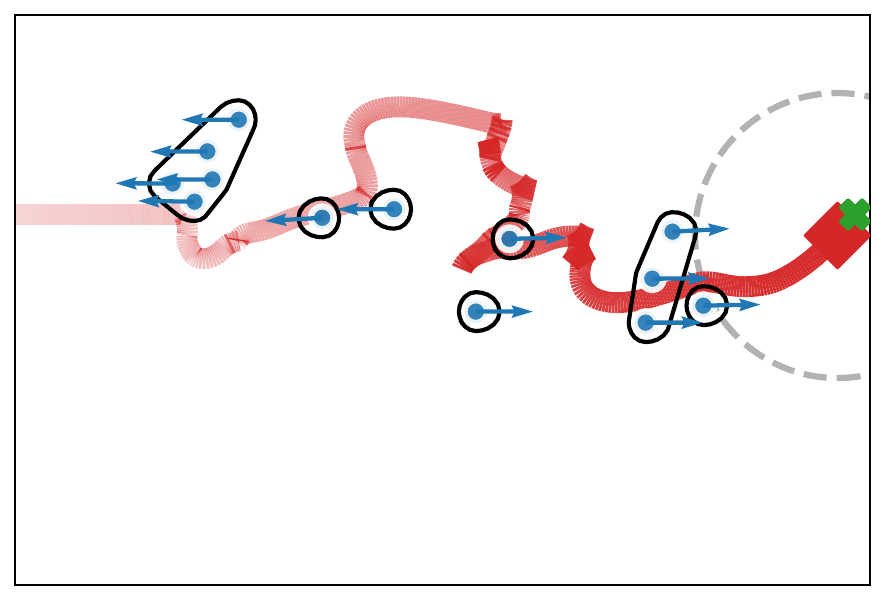}{t=63}%
\tile{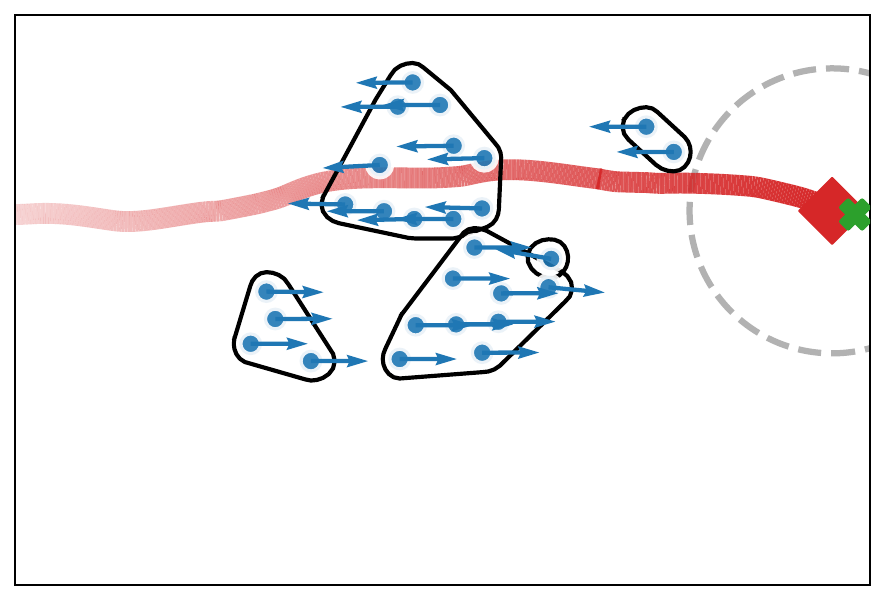}{t=33}%
\tile{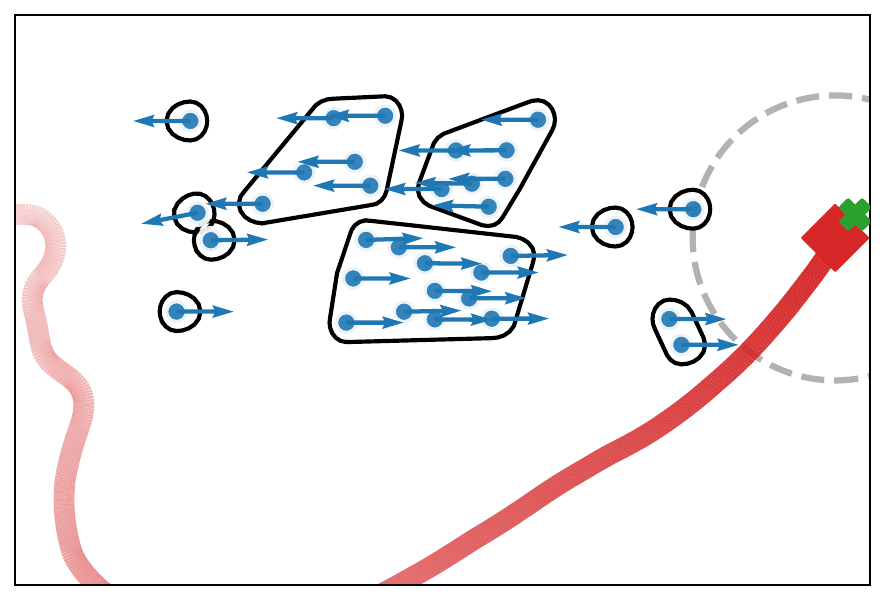}{t=52}\\

\caption{Example episode from the Synthetic dataset in the \textbf{online} setting ($t$ in seconds). The scenario contains two opposing pedestrian flows: one moving left to right and the other right to left. The robot starts on the left and aims for a goal on the right, located in the upper region where pedestrians move in the opposite direction. Taking the shortest direct path leads the robot into the dense opposing flow, which is initially outside the robot's observation radius. MPC and ORCA exhibit this behavior, becoming stuck in the oncoming flow and requiring much longer time to reach the goal. SARL avoids the dense crowd entirely, taking a long detour. CrowdAttn follows a straight path and later runs into the opposing flow, forcing the oncoming social groups to split apart for avoidance, resulting in multiple reactive actions and periods of freezing. In contrast, HiCrowd moves align with the pedestrian flow in the lower part of the scene, taking a detour but moving with the crowd, and reaches the goal in the shortest time.}
\label{fig:example_751}
\vspace{-2em}
\end{figure*}

\begin{figure*}
\centering
\includegraphics[width=0.85\linewidth]{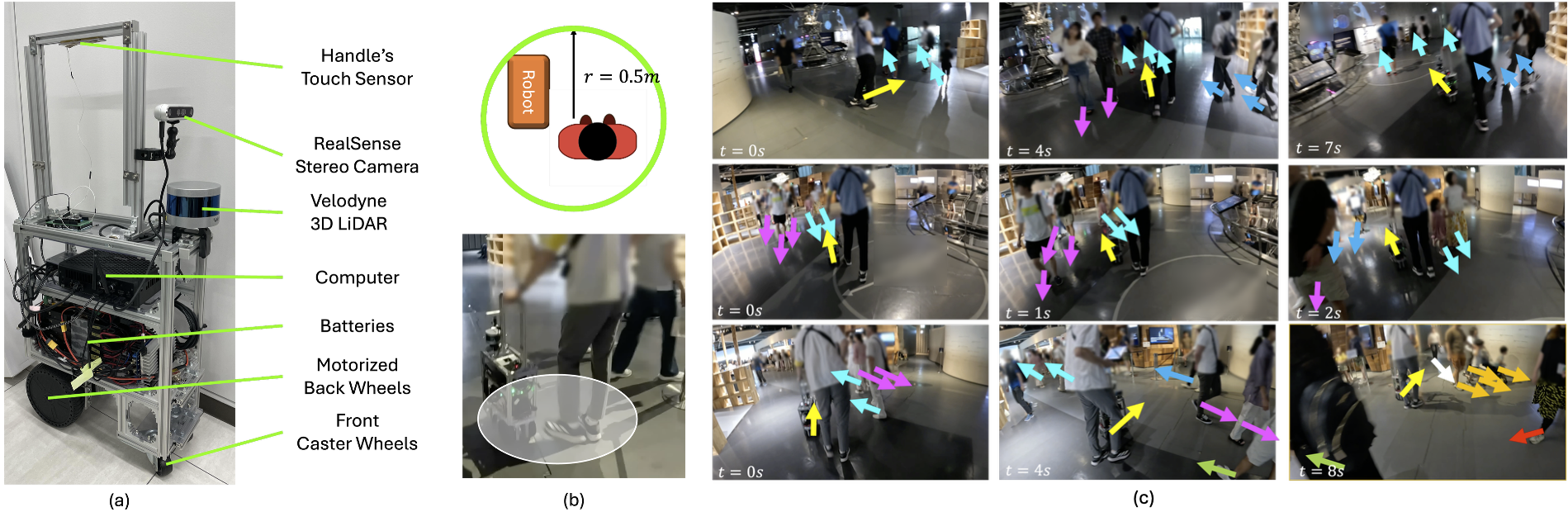}
\caption{Real-world experiment setup and examples. (a) Our differential-drive robot and its sensor configuration. (b) Illustration of the footprint of our robot. It is a circle of \SI{0.5}{\meter} radius that encompasses the robot and the predefined nominal human offset. (c) Qualitative examples. Robot movement is shown in yellow arrows. Different colors represent different groups of surrounding pedestrians. \textbf{Top row}: With the flow scenario. The robot merged into the pedestrian flow formed by the cyan and blue groups and naturally avoided the magenta group. \textbf{Middle row}: Against the flow scenario. The robot navigated successfully against the pedestrian flow, which consists of at least 3 groups. \textbf{Bottom row}: Crossing the flow scenario. The robot navigated through a crossing flow and reached to the goal successfully without collision.}
\label{fig:real-world}
\vspace{-1.5em}
\end{figure*}

\subsection{Quantitative Evaluation}
\begin{table}
    \centering
    \begin{tabular}{llllllll}
     \toprule
        \textbf{Method} & \textbf{SR$\uparrow$} & \textbf{CR$\downarrow$} & \textbf{TR$\downarrow$} & \textbf{NT$\downarrow$} & \textbf{PL$\downarrow$} & \textbf{MP$\uparrow$} & \textbf{FF$\downarrow$} \\
        \midrule
        \textbf{HiCrowd} & \textbf{1.00} & \textbf{0.00} & \textbf{0.00} & \textbf{19.10 }& \textbf{19.02} & 1.50 & \textbf{0.001} \\
        MPC & 0.98 & \textbf{0.00} & 0.02 & 28.29 & 27.78 & \textbf{2.01} & 0.010 \\
        CrowdAttn & \textbf{1.00} & \textbf{0.00} & \textbf{0.00} & 26.49 & 25.91 & 1.12 & 0.008 \\
        ORCA & 0.77 & \textbf{0.00} & 0.23 & 43.43 & 39.97 & 1.36 & 0.013 \\
        SARL & 0.78 & \textbf{0.00} & 0.22 & 33.25 & 25.90 & 1.19 & 0.051 \\
        \bottomrule
\end{tabular}
\caption{Navigation results in the \textbf{online} setting on the ETH-UCY dataset, where pedestrians react to the robot using ORCA simulation. Metrics include success rate (SR $\uparrow$), collision rate (CR $\downarrow$), timeout rate (TR $\downarrow$), navigation time (NT $\downarrow$), path length (PL $\downarrow$), minimum pedestrian distance (MP $\uparrow$), and freezing frequency (FF $\downarrow$).}
\label{tab:online_eth}
\vspace{-1em}
\end{table}

\begin{table}
    \centering
    \begin{tabular}{llllllll}
     \toprule
        \textbf{Method} & \textbf{SR$\uparrow$} & \textbf{CR$\downarrow$} & \textbf{TR$\downarrow$} & \textbf{NT$\downarrow$} & \textbf{PL$\downarrow$} & \textbf{MP$\uparrow$} & \textbf{FF$\downarrow$} \\
        \midrule
        \textbf{HiCrowd} & \textbf{1.00} & \textbf{0.00} & \textbf{0.00} & \textbf{30.88} & \textbf{30.85} & 1.35 & \textbf{0.000} \\
        MPC & \textbf{1.00} & \textbf{0.00} & \textbf{0.00} & 36.83 & 36.35 & 1.60 & 0.008 \\
        CrowdAttn & 0.96 & \textbf{0.00} & 0.04 & 45.45 & 43.22 & 1.45 & 0.017 \\
        ORCA & 0.97 & \textbf{0.00} & 0.03 & 40.89 & 39.85 & \textbf{1.61} & 0.002 \\
        SARL & 0.85 & \textbf{0.00} & 0.15 & 58.35 & 42.59 & 1.29 & 0.003 \\
        \bottomrule
\end{tabular}
\caption{Navigation results in the \textbf{online} setting on the Synthetic dataset.}
\label{tab:online_syn}
\vspace{-2em}
\end{table}
We evaluate our method HiCrowd against MPC, ORCA, SARL, and CrowdAttn in both \emph{online} (pedestrians react via ORCA) and \emph{offline} (pedestrians replay and do not react) settings on the ETH-UCY and Synthetic dataset. 

Evaluation results in the online setting are shown for ETH-UCY in \cref{tab:online_eth} and for Synthetic dataset in \cref{tab:online_syn}. HiCrowd achieves the best overall performance, with full success and the lowest navigation time and path length. In the online setting, baseline methods achieve high success rates but require longer navigation times, exhibiting more frequent reactive adjustments in dense crowd environments.

Evaluation results in the offline setting are shown for ETH-UCY in \cref{tab:offline_eth} and for Synthetic dataset in \cref{tab:offline_syn}.
When pedestrians do not react to the robot, RL-based baselines, including SARL and CrowdAttn, experience more collisions. The pure MPC baseline and ORCA require longer navigation times  and make more frequent reactive adjustments. HiCrowd maintains a higher success rate, which shows the benefit of aligning with  crowd flows rather than relying only on late reactive avoidance.

Notably, in the offline setting HiCrowd sometimes gets longer path lengths compared to baselines. These detours arise from following the pedestrian flow, which helps avoid potential future conflicts and collisions, leading to shorter navigation times and less freezing frequency. This illustrates the advantage of proactively aligning with social dynamics, where small path deviations ensure continuous and efficient progress toward the goal. Our method consistently achieves the lowest freezing frequency (FF) across all datasets and settings, showing that aligning with pedestrian flows enables continuous progress without getting stuck. In contrast, baselines can cause freezing situations, forcing the robot to stop.

\begin{table}
    \centering
    \begin{tabular}{llllllll}
     \toprule
        \textbf{Method} & \textbf{SR$\uparrow$} & \textbf{CR$\downarrow$} & \textbf{TR$\downarrow$} & \textbf{NT$\downarrow$} & \textbf{PL$\downarrow$} & \textbf{MP$\uparrow$} & \textbf{FF$\downarrow$} \\
        \midrule
        \textbf{HiCrowd} & \textbf{0.88} & 0.12 & \textbf{0.00} & \textbf{21.82} & \textbf{21.71} & 1.55 & \textbf{0.001} \\
        MPC & 0.76 & \textbf{0.09} & 0.15 & 40.65 & 36.53 & \textbf{2.11} & 0.056 \\
        CrowdAttn & 0.55 & 0.44 & 0.01 & 26.82 & 26.25 & 0.94 & 0.005 \\
        ORCA & 0.61 & 0.26 & 0.13 & 40.78 & 36.05 & 1.20 & 0.044 \\
        SARL & 0.24 & 0.71 & 0.05 & 29.73 & 25.67 & 0.76 & 0.014 \\
        \bottomrule
\end{tabular}
\caption{Navigation results in the \textbf{offline} setting on the ETH-UCY dataset, where pedestrians replay recorded trajectories and do not react to the robot.}
\label{tab:offline_eth}
\vspace{-1em}
\end{table}

\begin{table}
    \centering
    \begin{tabular}{llllllll}
     \toprule
        \textbf{Method} & \textbf{SR$\uparrow$} & \textbf{CR$\downarrow$} & \textbf{TR$\downarrow$} & \textbf{NT$\downarrow$} & \textbf{PL$\downarrow$} & \textbf{MP$\uparrow$} & \textbf{FF$\downarrow$} \\
        \midrule
        \textbf{HiCrowd} & \textbf{0.99} & \textbf{0.01} & \textbf{0.00} & \textbf{35.21} & 35.18 & \textbf{2.30} & \textbf{0.000} \\
        MPC & 0.91 & 0.09 & \textbf{0.00} & 36.08 & \textbf{34.93} & 1.92 & 0.023 \\
        CrowdAttn & 0.85 & 0.12 & 0.03 & 46.40 & 42.96 & 1.62 & 0.019 \\
        ORCA & 0.75 & 0.23 & 0.02 & 37.53 & 35.69 & 1.60 & 0.002 \\
        SARL & 0.45 & 0.50 & 0.05 & 57.57 & 42.68 & 1.33 & 0.040 \\
        \bottomrule
\end{tabular}
\caption{Navigation results in the \textbf{offline} setting on the Synthetic dataset.}
\label{tab:offline_syn}
\vspace{-1em}
\end{table}

On the other hand, although our method does not maximize the minimum pedestrian distance (MP) compared to baselines, it consistently maintains a social distance above \SI{1.2}{\meter}~\cite{hall1966proxemics} across all settings and datasets.

\subsection{Qualitative Evaluation}
\cref{fig:example_1904} and \cref{fig:example_751} show example trajectories for our method and all baselines in the ETH-UCY dataset (offline setting) and the Synthetic dataset (online setting), respectively. To visualize and compare complete trajectories, we allow all methods to continue until the end, even after collisions. In \cref{fig:example_1904}, which shows a dense crowd episode, baselines take longer paths and more reactive actions, and collisions happen with ORCA, SARL, and CrowdAttn. Our method, HiCrowd, reaches the goal efficiently without any collision. 
In \cref{fig:example_751}, two opposing pedestrian flows form a trap. Although the straight path to the goal is initially clear, it leads directly into the oncoming opposing flow, which lies beyond observation radius. In this situation, MPC and ORCA push into the opposing flow and freeze. CrowdAttn also takes the straight path and later runs into the opposing flow, forcing the oncoming social groups to split apart for avoidance, resulting in multiple reactive actions and periods of freezing. In contrast, our method detours to align with the left-to-right crowd flow and arrives in the shortest navigation time without collision. 

\vspace{-1mm}
\subsection{Ablation Study}
\begin{figure}
\centering
\includegraphics[width=.49\linewidth]{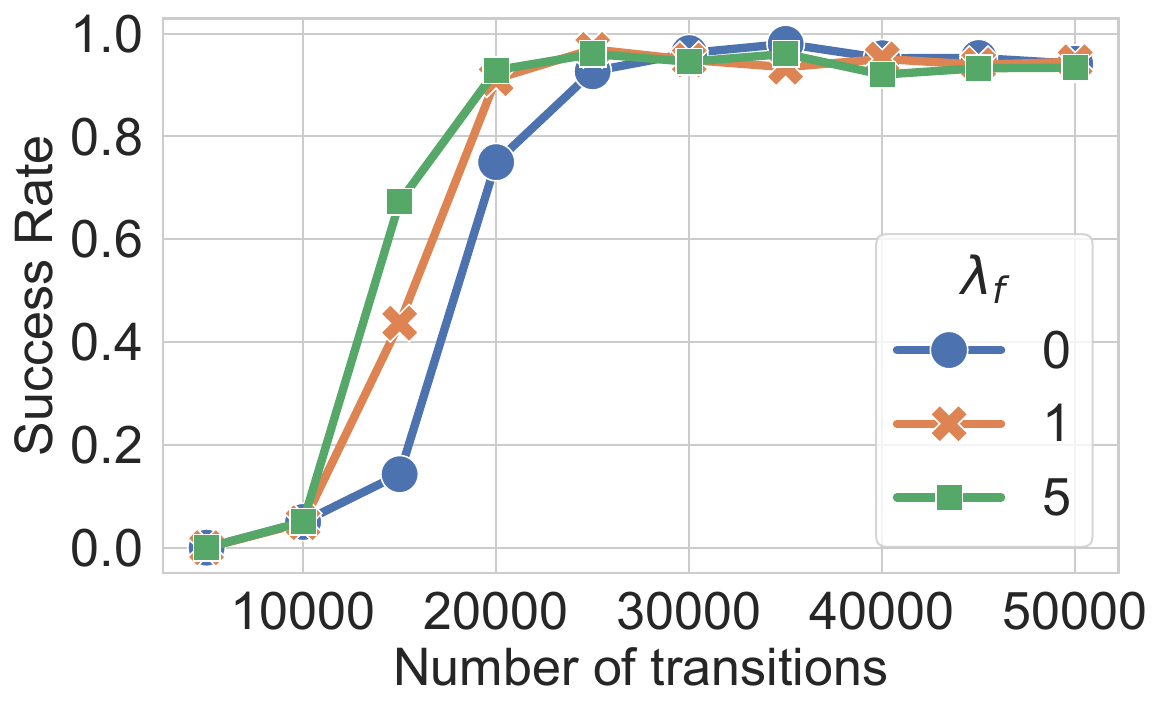}
\includegraphics[width=.49\linewidth]{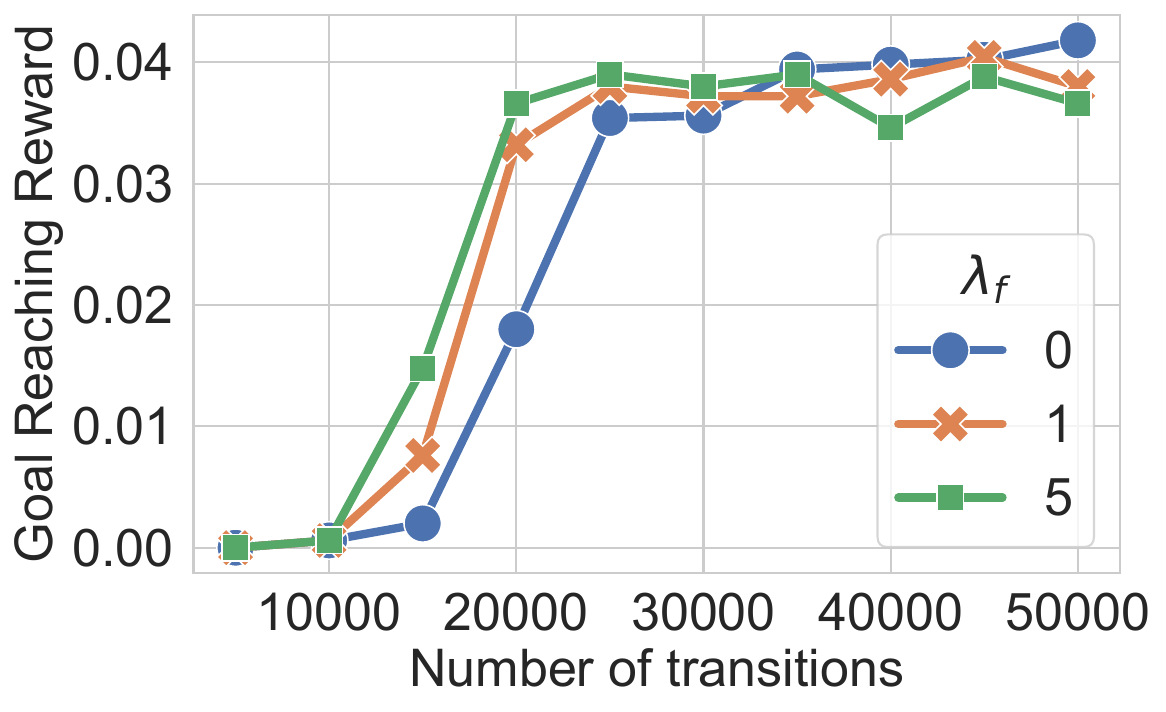}
\caption{Ablation on crowd following reward weight $\lambda_f$. Each transition corresponds to one RL policy decision step. Explicitly rewarding crowd following ($\lambda_f\in\{1,5\}$) accelerates learning and improves early performance, showing that following a moving crowd is an effective navigation strategy in dense environment.}
\label{fig:RL_curve}
\vspace*{-2em}
\end{figure}

To study the role of the crowd following reward $r_t^{\text{follow}}$, we perform an ablation with $\lambda_f \in \{0, 1, 5\}$ in \cref{fig:RL_curve}. When $\lambda_f>0$, the agent receives explicit guidance to follow the crowd, which accelerates learning and achieves higher performance with fewer transition steps. The faster convergence observed with higher $\lambda_f$ shows the usefulness of explicitly rewarding crowd following behavior, indicating that in dense environments, aligning with a moving crowd is an optimal navigation strategy. When $\lambda_f=0$, the agent relies only on goal-reaching rewards. In this case, learning is slower but still converges eventually, as crowd following behavior implicitly emerges as part of the optimal strategy in dense environments.

In addition, the MPC-only baseline serves as an ablation that removes the RL module and plans directly toward the goal.
Compared to HiCrowd, pure MPC provides safe short-horizon control but lacks long-term crowd reasoning, and can commit to locally clear paths that lead into opposing flows.
The RL module complements MPC by generating a follow point that guides the robot to align with suitable crowd flows, effectively abstracting long-term guidance beyond the immediate sensing horizon. This combination improves efficiency and reduces freezing, enabling efficient and safe navigation in dense crowds.

\subsection{Real world experiment}
To validate the effectiveness of HiCrowd in the real world, we deployed our model on the robot shown in Fig.~\ref{fig:real-world}-a. The robot is a custom platform built from scratch, referencing the open-source platform~\cite{guerreiro2019cabot}. We mounted a front-facing Intel RealSense D435 stereo camera and used the MMDetection toolbox~\cite{mmdetection} combined with depth images to obtain trajectories of the pedestrians in world coordinates. We also mounted a Velodyne VLP-16 light detection and ranging (LiDAR) sensor to localize the robot and safeguard against collisions. Due to the camera placement, the robot could not see the pedestrians to the sides or the rear. 
The robot's computer has an Intel Core i7-9750H CPU and a NVIDIA GeForce RTX 2070 GPU. The robot's speed is limited to \SI{1}{\meter\per\second} linear and $\tfrac{\pi}{4}\,\si{\radian\per\second}$.

We deployed the robot in a crowded real-world science museum during a holiday (Fig.~\ref{fig:real-world}-c). In line with our open-space assumption, we selected a wide area with few static obstacles. Our deployment was approved by our institution's ethics committee. To ensure the safety of the unsuspecting visitors, we attached a touch sensor to the top handle part of the robot, and the robot would only navigate autonomously when the handle was held\footnote{Experiments would not be permitted without this feature.}. We directly deployed HiCrowd trained in the simulator, and integrated it as a controller and planner plugin to the robot's ROS2-based software stack. The robot's footprint is a circle that encompasses both the robot and the nominal location of the human supervisor (Fig.~\ref{fig:real-world}-b).

We conducted 20 trials and tasked the robot to cross the selected museum area repeatedly. During the trials, the robot often needed to navigate around 10 people, and we did not observe any significant issues such as collisions or the freezing robot problem. In terms of computation, the RL policy estimated a follow point in $0.0047\pm0.0053$ seconds, and the time to produce an MPC action was $0.15\pm0.073$ seconds. Example robot behaviors are shown in Fig.~\ref{fig:real-world}-c. In the first example, the robot merged into the pedestrian flow formed by the cyan and blue groups and naturally avoided the magenta group. In the second example, the robot navigated successfully against the pedestrian flow characterized by at least 3 groups. In the third example, the robot navigated through a crossing flow defined by multiple groups. Please refer to our videos for more details.



\section{Conclusion}
\label{sec:conclusion}
We introduced HiCrowd, a hierarchical navigation framework that integrates RL and MPC to address the challenges of robot navigation in dense human environments. Rather than treating humans as dynamic obstacles~\cite{xie2023drlvo}, HiCrowd leverages surrounding pedestrian motion as guidance. A high-level RL policy generates a socially aware follow point that steers the robot toward a compatible crowd flow, while a low-level MPC controller tracks this guidance with safe, short-horizon planning. By encouraging alignment with nearby groups through a crowd-following reward, the framework improves both learning efficiency and navigation performance.
Extensive experiments across synthetic and real-world datasets demonstrate that HiCrowd outperforms classical and learning-based baselines, achieving higher success rates and more efficient navigation, also reducing freezing. The benefits are most pronounced in dense scenarios with structured human flows, where baseline methods can freeze or collide when encountering opposing crowds.

Finally, we deployed HiCrowd on a mobile robot and evaluated in two crowded real-world environments, a public museum and Expo 2025 Osaka, without any retraining. The robot consistently demonstrated socially compliant behaviors in scenarios of aligning with and crossing pedestrian flows, showing the real-world applicability of our approach.

The proposed method has limitations. First, the ratio between high-level and low-level actions is fixed, which is, the robot executes a number of MPC steps per follow point. This may lead to delayed reactions when crowd conditions change rapidly, or unstable behavior when follow points change too frequently. Future work will explore dynamically adapting the ratio based on flow variations or environmental uncertainty to improve responsiveness and stability. Static obstacles such as walls will also be explored. 
Lastly, the current framework treats groups primarily through their motion features without modeling their spatial extent. Incorporating group geometry could improve decision-making, particularly when navigating around large or irregularly shaped groups.










\printbibliography
\end{document}